\documentclass[conference]{IEEEtran}
\IEEEoverridecommandlockouts
\usepackage{cite}
\usepackage{amsmath,amssymb,amsfonts,multirow,}
\usepackage{graphicx}
\usepackage[table,xcdraw]{xcolor}
\usepackage{textcomp}
\usepackage{xcolor}
\usepackage{bm}
\usepackage{ctable}
\usepackage{caption, subcaption}
\usepackage{gensymb}
\usepackage{soul}
\usepackage{algorithm}
\usepackage{algpseudocode}
\usepackage{amsmath}
\usepackage{enumitem}

\algrenewcommand\algorithmicrequire{\textbf{Input:}}

\usepackage{authblk}

\def\BibTeX{{\rm B\kern-.05em{\sc i\kern-.025em b}\kern-.08em
 T\kern-.1667em\lower.7ex\hbox{E}\kern-.125emX}}
\begin{document}

\title{ED$^3$R: Energy-Aware Distributed Disaster Detection via Cooperative Agents in Robotic Systems}
\author{Lina Magoula*, Nikolaos Koursioumpas*, Nancy Alonistioti*, Ramin Khalili**
\\
* \emph{Dept. of Informatics and Telecommunications, National and Kapodistrian University of Athens, Greece} 
\\
** \emph{Huawei Heisenberg Research Center (Munich), Germany}
\\
\{lina-magoula*, nkoursioubas*, nancy*\}@di.uoa.gr \\
\{ramin.khalili**\}@huawei.com}

\maketitle

\begin{abstract}
Robotics are expected to support environmental monitoring and disaster detection, where decisions must be made under uncertainty, resource limitations, and strict operational constraints. In critical missions, such as wildfires, robots must not only identify hazardous events with sufficient confidence, but also manage the energy cost and time until detection. This paper introduces ED³R, an energy-aware distributed framework for wildfire detection under uncertainty that enables hierarchical cooperative decision-making between a robot and a remote controller. The remote controller decides upon the robot’s motion, while the robot senses the environment and decides where to execute the wildfire detection (onboard or remotely) and how. The common goal is to detect wildfires with a required confidence while minimizing the energy consumed by any robot operation. ED³R further integrates mechanisms to avoid nearby obstacles, prevent redundant exploration, enable adaptive early mission completion, and ensure feasibility through a custom penalty function. ED³R also introduces a forward-looking capability, enabled through distributed neural regression models that allow the agents to anticipate the future by evaluating candidate strategies before execution. The framework is evaluated through realistic robotics simulations, ablation studies, and baseline comparisons. ED³R achieves a mission success rate of up to 97.18\%, defined as the percentage of missions with true positive detections meeting the required confidence, excluding false positives and battery depletions. Especially in the most demanding missions, it reduces energy consumption by up to 36.4\% and detects wildfires up to 41\% faster than baselines.
\end{abstract}
\begingroup
\renewcommand{\abstractname}{Note to Practitioners}
\begin{abstract}
This work addresses a practical deployment question: how should an aerial robot allocate its limited battery while searching for a wildfire of unknown location and visibility? Results indicate that energy management should not be treated as a separate computing or communication problem. Motion, sensing, onboard processing, and data transmission are interdependent, and fixed policies can waste energy or delay detection. ED$^3$R coordinates the motion, processing, and communication decisions of an aerial robot and a remote controller during wildfire detection to achieve the required detection confidence while preserving battery energy. The evaluation provides four practical insights. First, ED$^3$R adapts effectively to different detection requirements, with a 75\% confidence threshold providing the best balance among energy consumption, detection performance, mission duration and success. Second, accurate forward-looking agent decisions result in higher mission success. Third, adaptive execution and cooperation are essential, since removing robot-side adaptation or coordination with the remote controller leads to inefficient strategies. Fourth, ED$^3$R outperforms centralized neural-based, heuristic and Reinforcement Learning baselines in reliability, energy efficiency, and speed. It achieves up to 97.18\% mission success and, in the most demanding missions, reduces energy consumption by up to 36.4\% and detects wildfires up to 41\% faster than the baselines. The results are based on realistic robotic simulations, and future extensions will consider multi-robot cooperation.
\end{abstract}
\endgroup


\section{Introduction}\label{Intro}

Over the past decades, robotics has emerged as a transformative technology, finding applications in fields ranging from manufacturing to healthcare \cite{shukla2025historical}. However, its impact extends beyond technological innovation, reaching a broader scope shaped by societal and environmental \textit{values} and guided by the Sustainable Development Goals (SDGs) \cite{un2015transforming}. In this context, robotics can support SDGs such as climate action, sustainable cities and communities, and life on land by enabling safer, faster, and more data-driven interaction in complex environments \cite{articleFilho}.

Sustainability-oriented robotics is particularly relevant in environmental monitoring, where early, accurate, and continuous observation is essential for reducing risks to human lives, infrastructure, and ecosystems \cite{Ghassemian2026}. The escalating impact of natural disasters, such as floods and wildfires, highlights the pressing demand for scalable and adaptive monitoring solutions capable of addressing the spatial scale, operational complexity, and dynamicity of such environmental crises. This necessity is particularly evident in the context of wildfires, for which the European Union Agency for the Space Programme (EUSPA) defines the requirement for early detection to be achieved within 10 minutes of ignition\cite{EUSPA2024EMAidUserNeeds}. Consequently, robotic systems operating under this \textit{uncertainty} should be equipped with self-learning capabilities and context-aware autonomy, enabling them to adapt their behavior rapidly in response to ever-changing conditions and time-critical operational demands. This requirement motivates the transition toward advanced forms of intelligence, such as agentic intelligence, where robotic agents are not limited to predefined actions but are capable of perception, reasoning, decision-making, execution, and communication according to environmental and operational context \cite{one6G2023, 6G-IA_2024_EuropeanVision}. 

However, in real-world scenarios, such agentic capabilities are strongly coupled with the availability of computational, communication and energy resources, introducing significant resource related demands. Constrained onboard computing capabilities can restrict perception, reasoning, and control, while fluctuating network conditions can limit communication with other agents or remote infrastructure. Most importantly, for every movement, sensing, computation, and communication a robot consumes energy, directly affecting its autonomy, mobility range and duration. Therefore, agentic robotic systems must not only be intelligent and context-aware, but also resource-aware and energy-efficient. From processing environmental sensing information to selecting actions, coordinating with other agents, and communicating decisions, robots must balance performance against energy consumption.

To address the need for integrating agentic robotic systems into natural crisis management, this study introduces ED$^3$R, an energy-aware distributed framework for wildfire detection under \textit{uncertainty}. Unlike conventional robotic monitoring approaches that primarily focus on detection performance, ED$^3$R jointly considers mission effectiveness and energy efficiency by requiring wildfire detection with a target confidence while minimizing the energy consumed by robot operations (mobility, sensing, computation, and communication). ED$^3$R enables hierarchical cooperative decision-making between a robot and a remote controller ($\text{RC}$), allowing the two agents to make distributed yet coordinated decisions.

A pool of AI wildfire detection models, differing in complexity and performance, is available to the agents. The $\text{RC}$ agent is responsible for \textit{motion-control} decisions, while the robot agent senses the environment and makes resource-aware decisions on both \textit{where} wildfire detection should be executed and \textit{how} it should be performed. The \textit{where} decision selects between onboard and remote execution, while the \textit{how} decision selects the most suitable AI model from the available pool. In the case of remote execution, the robot transmits its sensing data to the $\text{RC}$, which \textit{assists} the mission through remote wildfire detection.

ED$^3$R follows a two phase methodology that combines offline knowledge acquisition with online real-time decision-making. In the offline phase, both agents explore the environment and learn how their actions affect wildfire detection confidence, energy consumption, and mission effectiveness. In the online phase, this acquired knowledge is used to support real-time distributed and interdependent decisions during mission execution. 

The methodology is supported by a variety of mechanisms that improve safety, efficiency, and reliability. A 360$\degree$-obstacle avoidance mechanism prevents unsafe robot motion near obstacles. A region avoidance mechanism reduces redundant exploration and redirects the robot toward more promising or unexplored regions. An adaptive early stopping mechanism terminates the mission once reliable wildfire detection is sustained, avoiding unnecessary motion and energy consumption. In addition, a custom penalty function promotes feasible strategies that respect system constraints. 

Another key innovation of ED$^3$R is its \textit{forward-looking} decision-making capability. Since the $\text{RC}$ and the robot act in a distributed manner but share a common mission reward only after their actions are completed, each agent must \textit{look forward} by estimating the shared mission outcome of its candidate decisions, while the $\text{RC}$ must also account for the robot’s subsequent actions. ED$^3$R enables this through distributed neural regression models to evaluate multiple candidate strategies before action execution. The agents then greedily select the strategy expected to provide the best trade-off between detection confidence and energy efficiency.
The key contributions can be summarized as follows:
\begin{itemize}
\item A distributed hierarchical cooperative decision-making framework between two agents that jointly optimizes mission effectiveness and energy efficiency.
\item A greedy \textit{forward-looking} mechanism that evaluates possible futures by anticipating their consequences prior to action selection.
\item The design and integration of robotics-oriented mechanisms that enhance safety, efficiency, and reliability in mission-critical operations.
\item Formulations of energy consumption and communication networks, along with computer vision models, based on realistic state-of-the-art literature.
\item A real-time evaluation process using state-of-the-art robotics simulations. 
\item Robust performance, in terms of detection confidence, mission duration (flight time until reliable wildfire detection), energy consumption and overall mission success, even under unseen wildfire incidents and wireless network setups.
\item Comprehensive ablation studies and baseline comparisons that demonstrate the effectiveness and advantages of our proposed approach.

\end{itemize}

The rest of the paper is organized as follows. Section \ref{related_work} presents relevant state-of-the-art literature. Section \ref{system_model} provides the system model and section \ref{problem_formulation} provides the problem formulation. The proposed solution is presented in Section \ref{proposed_algorithm} and evaluated in Section \ref{performance_evaluation}. Finally, Section \ref{conclusions} concludes the paper. Additional details related to the problem formulation and performance evaluation are provided in Appendix \ref{appendix}.

\section{Related Work}\label{related_work}
This section presents an overview of the state-of-the-art. These efforts tackle different objectives related to energy efficiency, search and rescue (SAR) task planning and time-critical applications in robotic systems.

\textbf{Energy consumption}: 
Energy-aware optimization has been studied in cooperative robotic systems. Existing works optimize different aspects of robotic operation, including navigation, sensing, infrastructure resource allocation \cite{10142027Romero}, computation offloading and latency-aware resource management \cite{9762674Yin,drones8100564}, distributed task allocation for transportation \cite{Djenadi02012022}, and reinforcement learning (RL)-based post-disaster exploration \cite{XUE2026109171}. These approaches primarily aim to reduce energy consumption, balance robot workloads, or satisfy latency constraints. 

\textbf{Search and Rescue (SAR)}: 

Several works investigate UAV-based optimization for hazardous SAR scenarios. The work \cite{Han8040138} addresses rescue-task allocation by minimizing UAV deployment, rescue cost, and operation duration through a hybrid optimization algorithm. Work \cite{Farsath10580372} focuses on dynamic hazardous environments, combining image recognition and object detection with RL for adaptive navigation and path optimization. In \cite{Horyna101007} a decentralized UAV swarm system targets autonomous exploration and object detection without explicit communication. In \cite{XING2022102972} a cooperative multi-UAV system is proposed that integrates real-time target detection, digital elevation model-based path planning, and monocular-camera-based positioning for GPS-level target localization.

\textbf{Miscellaneous tasks}: 
There is also a number of works focusing on tasks related to task planning and time-critical applications. In \cite{Urbaniak2024}, image processing is offloaded to evaluate its impact on reaction time in collaborative robotics. In \cite{9676458Zhou}, a Graph Neural Network (GNN) is deployed to improve inference accuracy and robustness to sensing failures. Works \cite{damigos2024,10002890Cui} propose network- and resource-aware coordination for time-critical operation in dynamic robotic networks. 

\textbf{Wildfire Detection}: A comparison of the most relevant wildfire detection and monitoring studies is presented in Table \ref{tab:sota-table}, highlighting their key aspects. Robotic wildfire monitoring
has been investigated from several complementary perspectives. Distributed control approaches coordinate multiple UAVs for fire-front tracking
\cite{8331947Pham}, while multi-objective evolutionary methods optimize wildfire coverage and operational cost \cite{9504947Shrestha}. Persistent
monitoring and learning-based studies support fire observation and stochastic propagation over a time horizon \cite{Patrinopoulou2024, Julian2019, Seraj2022}. Energy-aware UAV perception systems jointly adapts image resolution, flight and detection parameters to reduce the energy cost of aerial object detection \cite{11315192Akpomedaye, 10206033Suo}.

Table~\ref{tab:sota-table} compares ED$^3$R with representative wildfire detection and monitoring methods in terms of different aspects. Existing energy- and perception-aware works typically optimize individual aspects or subsets of robot operation (e.g., navigation, offloading, sensing or onboard perception) and do not jointly coordinate motion, computation, and communication. Tracking and coverage methods focus mainly on motion and resource allocation, often assuming fire observations are available. Distributed neural decision-making and forward-looking are also examined in a few studies, mainly through RL-based policies. However, independent distributed decision-making would not capture the interdependencies among agents, since each agent would optimize its own decision without accounting for its effect on subsequent actions of others. This gap motivates ED$^3$R, which jointly considers these interdependent decisions within a unified distributed hierarchical cooperative framework under time-varying communication and computational resources in dynamic wildfire scenarios. Through the introduced hierarchy and a shared mission objective, ED$^3$R coordinates the agents, while its forward-looking capability allows each agent to anticipate the effect of its own decisions on the subsequent decisions of the others.

\begin{table}[!ht]
\centering
\resizebox{\columnwidth}{!}{%
\begin{tabular}{cccccc}
\hline
\textbf{Studies} & \textbf{\begin{tabular}[c]{@{}c@{}}Energy \\ Efficiency\end{tabular}} & \textbf{\begin{tabular}[c]{@{}c@{}}Computer \\ Vision\end{tabular}} & \textbf{\begin{tabular}[c]{@{}c@{}}Joint Control: \\ Motion, \\ Computation,\\ Communication\end{tabular}} & \textbf{\begin{tabular}[c]{@{}c@{}}Distributed NN\\ Decision-Making\end{tabular}} & \textbf{\begin{tabular}[c]{@{}c@{}}Forward\\ Looking\end{tabular}} \\ \hline
\rowcolor[HTML]{EFEFEF} 
\cite{8331947Pham} & \textbf{x} & \textbf{x} & motion & distributed, not NN & \textbf{x} \\
\rowcolor[HTML]{FFFFFF} 
\cite{9504947Shrestha} & \checkmark & \textbf{x} & motion & evolutionary & \textbf{x} \\
\rowcolor[HTML]{EFEFEF} 
\cite{Patrinopoulou2024} & \textbf{x} & \textbf{x} & motion & distributed, not NN & \textbf{x} \\
\rowcolor[HTML]{FFFFFF} 
\cite{11315192Akpomedaye} & \checkmark & \checkmark & \textbf{x} & \textbf{x} & \textbf{x} \\
\rowcolor[HTML]{EFEFEF} 
\cite{10206033Suo} & \checkmark & \checkmark & no communication & \textbf{x} & \textbf{x} \\
\rowcolor[HTML]{FFFFFF} 
\cite{Seraj2022} & \textbf{x} & \textbf{x} & motion & \textbf{x} & \checkmark \\
\rowcolor[HTML]{EFEFEF} 
\cite{Julian2019} & \textbf{x} & \textbf{x} & motion & \checkmark & \checkmark \\
\rowcolor[HTML]{FFFFFF} 
\cellcolor[HTML]{FFFFFF}\cite{khoshdel2024pyrotrack} & \checkmark & \textbf{x} & motion & \textbf{x} & \checkmark \\
\rowcolor[HTML]{EFEFEF} 
{\color[HTML]{000000} \cite{drones9010017}} & {\color[HTML]{000000} \textbf{x}} & {\color[HTML]{000000} \textbf{x}} & {\color[HTML]{000000} motion} & {\color[HTML]{000000} \textbf{x}} & {\color[HTML]{000000} \textbf{x}} \\
\rowcolor[HTML]{FFFFFF} 
\cellcolor[HTML]{FFFFFF}\cite{11005691Diaz} & \checkmark & \textbf{x} & \begin{tabular}[c]{@{}c@{}}motion \&\\ communication\end{tabular} & \checkmark & \checkmark \\
\rowcolor[HTML]{EFEFEF} 
\cite{10834526Diaz} & \checkmark & \textbf{x} & \begin{tabular}[c]{@{}c@{}}motion \&\\ communication\end{tabular} & \checkmark & \checkmark \\
\rowcolor[HTML]{FFFFFF} 
\cellcolor[HTML]{FFFFFF}\cite{10838709Mu} & \textbf{x} & \checkmark & computation & \textbf{x} & \textbf{x} \\
\rowcolor[HTML]{C0C0C0} 
\textbf{ED$^3$R} & \checkmark & \checkmark & \checkmark & \checkmark & \checkmark \\ \hline
\end{tabular}%
}
\caption{State-of-the-art Comparison Table: Wildfire Detection \& Disaster Management}
\label{tab:sota-table}
\end{table}

\section{System Model} \label{system_model}
This section presents the system model of two agents, a single UAV robot and a remote controller, both equipped with Artificial Intelligence (AI) capabilities, tasked with a mission to cooperatively explore an unknown environment for potential wildfire incidents. The robot is capable of sensing the environment and performing onboard wildfire detection, while the remote controller, denoted by $\text{RC}$, handles the robot’s motion selection and is also equipped with a wildfire detection algorithm $l_{\mathrm{RC}}$ to assist the robot when needed.

We consider time is divided into discrete time units named as timesteps. At each timestep $k \in \mathbb{N}$, the $\text{RC}$ selects the robot's trajectory and pose represented by $\mathcal{P}_{k}$. For simplicity, trajectory and pose selection will be referred to as motion command. The time required for the robot to take the motion command varies and is denoted by $t_{k} \in \mathbb{R}_+$. The robot is powered by an onboard battery with a maximum energy capacity $\beta^{\max} \in \mathbb{R}_+$. Let $\beta_{k} \in \mathbb{R}_+$ denote the robot's available battery capacity at timestep $k$. The robot consumes energy to offer a series of capabilities:

\textbf{Robot Movement Capabilities}: The robot can move along the three axes, rotate 360$\degree$, and is able to actuate its joints to control the orientation of its onboard sensors (e.g. camera gimbal). Let $\{x_k,y_k,z_k\}$ denote the robot movement along axes, $\mathtt{r}_{k}$ denote the robot's body rotation and $\textbf{so}_k$ the orientation of each sensor at $k$.

\textbf{Robot Sensing Capabilities}: At each timestep, the robot senses the environment for a fixed time period denoted by $\mathtt{s} \in \mathbb{R}_+$.\footnote{We assume that $\mathtt{s}$ is sufficient to acquire newly generated sensing data.} Let $\mathcal{S}$ denote the set of sensors available at the robot. Let $I_s$ denote the sensing sample rate of sensor $s \in \mathcal{S}$ and $w_{s}$ the size of each data sample. 

\textbf{Robot Computational Capabilities}: We assume time-varying computational resources at the robot. As such, let $f_{k} \in \mathbb{R}_+, n^{\mathtt{cpu}}_{k} \in \mathbb{N}$ denote the available computational capacity (i.e., CPU speed) and number of CPU cores available at timestep $k$, and $\varsigma \in \mathbb{R}_+$ the CPU's effective switched capacitance (ESC). At each timestep, the robot can complete a certain number of Floating Point Operations (FLOPs) per cycle, denoted by $c_{k} \in \mathbb{R}_+$. 

\textbf{Robot Perception Capabilities}: The robot perceives the environment to perform wildfire detection. A set of AI algorithms  
$L$ is available for the robot to select at each $k$. The computational complexity $\alpha_{l_{k}} \in \mathbb{R}_+$ of a selected AI algorithm $l_{k} \in L$, depends on the architecture and input size, and is measured in FLOPs. The average performance of an AI algorithm at $k$ is denoted by $\eta_{l_{k}}$ and could refer to metrics such as confidence score and accuracy. The time required by the robot to process one sample using $l_{k}$ is denoted by $\tau_{\alpha_{l_{k}}} \in \mathbb{R}_+$. 

\textbf{Robot Communication Capabilities}: At each $k$, the robot may request assistance from the $\text{RC}$ by offloading its sensing data. Then, the $\text{RC}$ executes $l_{\mathrm{RC}}$ and achieves an average performance $\eta_{l_{\mathrm{RC},k}}$. The communication conditions between the robot and the $\text{RC}$ are time-varying and the channel is modeled as a flat-fading with Gaussian noise power density $N_0 \in \mathbb{R}$ and channel gain $g_{k} \in \mathbb{R}$, where the fading is assumed constant during the timestep. Also, let $b_{k} \in \mathbb{R}_+$ and $p_{k} \in \mathbb{R}_+$ be the available bandwidth and transmission power of the robot, respectively. By $\mathtt{d}_{k} \in \mathbb{R}_+$ we denote the achievable data rate of the communication channel. 

Table \ref{table_notations} summarizes all the notations of the system model, while problem formulation and optimization-specific variables are introduced in Section \ref{problem_formulation}.

\begin{table}[!ht]
\centering
\resizebox{\columnwidth}{!}{%
 \begin{tabular}{|p{0.065\textwidth}|p{0.4\textwidth}|}
\hline
\cellcolor[HTML]{EFEFEF}\textbf{Notation} & \cellcolor[HTML]{EFEFEF}\textbf{Description} \\
\hline
$\text{RC}$ & The AI-enabled remote controller \\ \hline
$k$ & The $k^{th}$ timestep\\ \hline
$\mathcal{P}_{k}$ & The motion command at $k$\\ \hline
$t_{k}$ & The time required for the robot to take an assigned pose at $k$\\ \hline
$\mathtt{s}$ & The sensing period for the robot \\ \hline 
$\beta^{\max}$ & The robot's maximum battery capacity \\ \hline
$\beta_{k}$ & The robot's available battery capacity at $k$ \\ \hline
$f_{k}$ & The robot's available computational capacity at $k$ \\ \hline
$n^{\mathrm{cpu}}_{k}$ & The number of robot's available CPU cores at $k$ \\ \hline
$c_{k}$ & The number of FLOPs per cycle the robot can complete at $k$ \\ \hline
$\varsigma$ & The robot's CPU effective switched capacitance \\ \hline
$\mathcal{S}$ & The set of installed robot's sensors \\ \hline
$I_s$ & The sensing sample rate of robot's sensor $s$ \\ \hline
$w_{s}$ & The data sample size of robot's sensor $s$ \\ \hline
$L$ & The set of AI algorithms available at the robot \\ \hline
$l_{k}$ & The selected AI algorithm at $k$ \\ \hline
$\alpha_{l_{k}}$ & The computational complexity of the AI algorithm $l_{k}$ at $k$\\ \hline
$l_{\mathrm{RC}}$ & The AI algorithm at the remote controller $\text{RC}$ \\ \hline
$\alpha_{l_{\mathrm{RC}, k}}$ & The computational complexity of the $\text{RC}$'s $l_{\mathrm{RC}}$ at $k$\\ \hline
$\eta_{l_{k}}$ & The average performance of $l_k$ executed at $k$ \\ \hline
$\eta_{l_{\mathrm{RC},k}}$ & The average performance of $l_{\mathrm{RC}}$ executed at $k$\\ \hline
$\tau_{\alpha_{l_{k}}}$ & The computation time to execute $l_{k}$ at $k$ \\ \hline
$\mathtt{d}_{k}$ & The robot's achievable transmission data rate at $k$ \\ \hline
$N_0$ & The white Gaussian noise power spectral density\ \\ \hline
$g_{k}$ & The gain of the wireless channel the robot has access to at $k$ \\ \hline
$b_{k}$ & The robot's allocated bandwidth at $k$ \\ \hline
$p_{k}$ & The transmission power of the robot at $k$ \\ \hline
\end{tabular}
}
\caption{System Model Notation Table} \label{table_notations} 
\vspace{-10pt}
\end{table}

\section{Problem Formulation}\label{problem_formulation}
Both the $\text{RC}$ and the robot are tasked to cooperatively detect a wildfire incident with a certain confidence. Satisfying this confidence level alone is not practical, since each robot operation, including movement, sensing, computation, and communication, consumes energy. As a result, the robot may over-consume energy to detect a single wildfire incident, thereby limiting its ability to continue surveying the environment. Therefore, to ensure mission success, we treat detection confidence as a constraint and minimize the energy required by the robot to satisfy this confidence level. This allows the robot to achieve reliable detection while preserving energy for continuous operation.

Achieving this objective requires a chain of hierarchical, distributed, and interdependent decisions taken by both the $\text{RC}$ and the robot. Each agent controls complementary parameters and operates under a different view of the same environment, necessitating coordinated decision-making.

\textbf{Decision-Making Stage 1 ($\text{RC}$)}: At the beginning of each $k$, the first stage of decision-making takes place at the $\text{RC}$. The $\text{RC}$ decides upon a motion command ($\mathcal{P}_{k}$). The assigned command will influence all subsequent robot's decisions, as motion precision affects the robot's perspective and can facilitate progress toward achieving the end goal and vice versa.

\subsection{Energy Requirements for Robot Movement} \label{energy_movements}

To follow the $\text{RC}$'s motion command, the robot executes multiple movements. Three motion types are considered: hovering, horizontal, and vertical movement, each with distinct energy demands. The corresponding energy models are explicitly derived from a variety of previously published and experimentally grounded studies, selected for their alignment with real-world UAV applications \cite{Thibbotuwawa101007,Pollet03832135,Dorling7513397,Marins8691921,Muli101007}. 

The total energy consumption for the robot movement at $k$, denoted by $E_{\mathrm{movement}}(t_{k})$, is expressed as:
\begin{equation}
    E_{\mathrm{movement}}(t_{k})  = \sum_{\mathtt{m} \in \mathcal{M}} P_\mathtt{m}\cdot t_{\mathtt{m,k}}  \quad(Joules),
\end{equation}
where:
\vspace{-6pt}
\begin{equation}\label{movement_time}
   \sum_{\mathtt{m} \in \mathcal{M}} t_{\mathtt{m,k}} =  t_{k}.
\end{equation}
$\mathtt{m} \in \mathcal{M} = \{hovering, horizontal, vertical\}$ represents the set of movement types, $P_\mathtt{m}$ is the power of each $\mathtt{m}$ ($Watts$), and $t_{\mathtt{m,k}}$ is the time spent on each $\mathtt{m}$ ($secs$) at $k$. 
The power $P_{\mathtt{m}}$ of each movement type is provided in the Appendix.

\textbf{Decision-Making Stage 2 (Robot):} Following the motion command (at $t_k$) and before sensing the environment, the robot decides on a binary variable $o_{k}$ indicating whether to offload the detection task or proceed with local execution. In the case of local execution, the robot also decides on an appropriate AI algorithm $l_{k}$.

\begin{figure}[ht!]
 \centering
 \includegraphics[width=0.9 \linewidth]{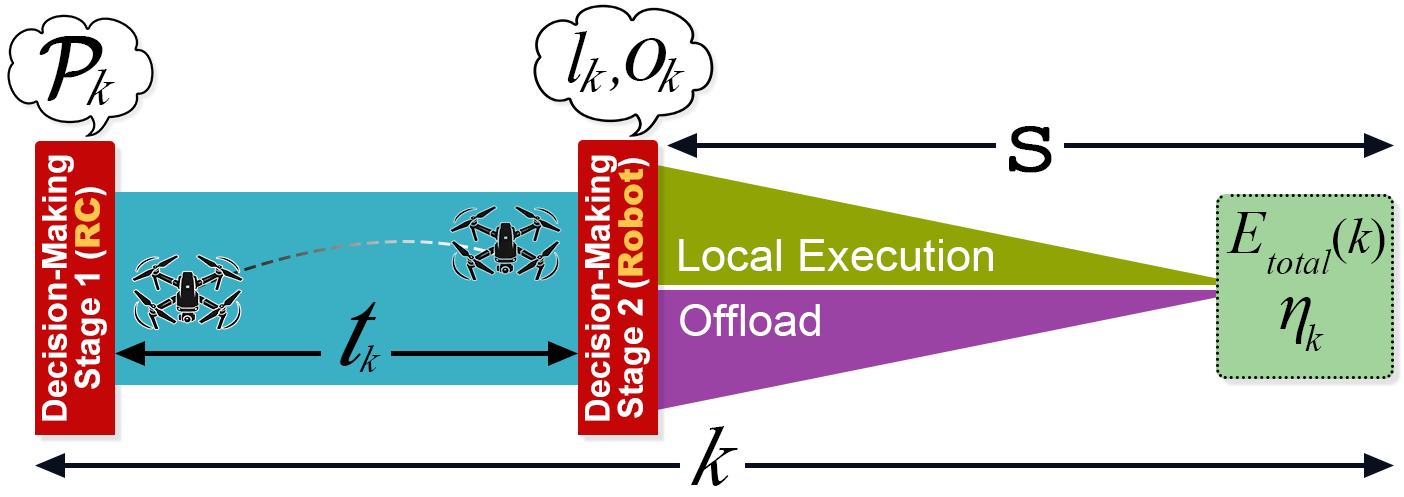}
 \caption{Timeline of timestep $k$ - Decision-Making Stages}
 \label{timeline_decisions}
\end{figure}

Figure \ref{timeline_decisions} illustrates the timeline of timestep $k$, showing the chain of decisions, their dependencies, and when each decision is made. 

\vspace{-7pt}
\subsection{Energy Requirements for Robot Sensing}\label{energy_sensing}

The robot senses the environment for a period $\mathtt{s}$ using its set of sensors $\mathcal{S}$. The total energy required for sensing at $k$, denoted by $E_{\mathrm{sensing}}(\mathtt{s})$, is expressed as:

\begin{equation}
    E_{\mathrm{sensing}}(\mathtt{s}) = \sum_{s \in \mathcal{S}} P_{s} \cdot \mathtt{s} \quad ({Joules}),
\end{equation}
\vspace{-6pt}
where:
\begin{equation}
    P_{s} = I_s \cdot w_{s} \cdot E_{(bit, s)} \quad ({Watts}).
\end{equation}
$P_{s}$ is the power of sensor $s$ ($Watts$) and is based on its type (e.g. Light Detection And Ranging (LiDAR), optical camera, Global Positioning System (GPS)), sensing rate and data sample size. $E_{(bit,s)}$ denotes the energy that sensor $s$ consumes to process one $bit$.

\subsection{Energy Requirements for Local Execution} \label{local_energy_detection}

In case of local execution, the robot processes the sensing data locally using the selected AI algorithm $l_k$. The computation time and energy consumption required for executing $l_k$, assuming perfect parallelization across $n^{\mathtt{cpu}}_{k}$ cores, is denoted by $T_{\mathrm{computation}}(\alpha_{l_{k}})$, $E_{\mathrm{computation}}(\alpha_{l_{k}})$ and are given by:
\begin{equation}
\begin{aligned}
    T_{\mathrm{computation}}(\alpha_{l_{k}}) = 
     \frac{\mathtt{N}(I_{s}, \mathtt{s}) \cdot\alpha_{l_{k}}} {c_{k} \cdot n^{\mathtt{cpu}}_{k} \cdot f_{k}} \quad (seconds),
\end{aligned}
\end{equation}
\vspace{-6pt}
\begin{equation}
\begin{aligned}
    E_{\mathrm{computation}}(\alpha_{l_{k}}) = 
     \frac{\varsigma \cdot \mathtt{N}(I_{s}, \mathtt{s}) \cdot \alpha_{l_{k}} \cdot f_{k}^2} {c_{k}} \quad (Joules),
\end{aligned}
\end{equation}
where: 
\vspace{-5pt}
\begin{equation}
\begin{aligned}
    \mathtt{N}(I_{s}, \mathtt{s}) = \sum_{s \in \mathcal{S}} I_{s} \cdot \mathtt{s},
\end{aligned}
\end{equation}
denotes the number of data samples collected during the sensing period $\mathtt{s}$ to be processed by $l_{k}$.

\subsection{Energy Requirements for Offloading} \label{remote_energy_detection}
In case of offloading, the robot transmits the sensing data for remote wildfire detection. The time and energy required for transmission, denoted by $T_{\mathrm{transmission}}(k)$, $E_{\mathrm{transmission}}(k)$, are:
\vspace{-6pt}
\begin{equation} \label{transmission_energy}
  T_{\mathrm{transmission}}(k) = \frac{\sum_{s \in \mathcal{S}}\mathtt{N}(I_{s}, \mathtt{s}) \cdot w_{s}}{\mathtt{d}_{k}} \quad (seconds),
\end{equation}
\vspace{-10pt}
\begin{equation} \label{transmission_energy}
  E_{\mathrm{transmission}}(k) = p_{k} \cdot  T_{\mathrm{transmission}}(k) \quad (Joules),
\end{equation}
where the modeling of the achievable data rate at $k$ is provided in the Appendix.

\subsection{Detection Confidence} \label{ai_performance_requirements}
The robot senses the environment over a time window $\mathtt{s}$ and performs wildfire detection. The detection confidence level may fluctuate over time due to variations in sensing quality, introducing detection uncertainty. To capture this, we introduce a temporal consistency measure, named the Stable Confidence Score (SCS), to quantify detection stability over $\mathtt{s}$. The SCS denotes the confidence level obtained at the end of $k$ by performing either local detection, $\eta_{l_k}$, or remote detection, $\eta_{l_{\mathrm{RC},k}}$. This measure is denoted by $\eta_k$ and is defined as:
\vspace{-5pt}
\begin{equation}
    \eta_{k} = \max\Big(0, \;\overline{\mathrm{cs}_{l_k}} \cdot (1 - \frac{\sigma_{{\mathrm{cs}_{l_k}}}}{\overline{\mathrm{cs}_{l_k}}})\Big), 
\end{equation}
where $\overline{\mathrm{cs}_{l_k}}$ denotes the average confidence of algorithm $l_k$ over $\mathtt{s}$, and $\sigma_{\mathrm{cs}_{l_k}}$ represents the corresponding standard deviation. The ratio $\frac{\sigma_{\mathrm{cs}_{l_k}}}{\overline{\mathrm{cs}_{l_k}}}$ is the coefficient of variation, measuring dispersion around the mean. 

\subsection{Problem Objective}
Overall, this chain of decisions will lead at the end of $k$ to a total energy consumption $E_{\mathrm{total}}(k)$ and detection confidence $\eta_{k}$ that contributes to the overall objective given below:
\begin{equation} \label{overall_objective}
\begin{split}
\min_{\mathcal{P}_{k},\, o_{k},\, l_{k}} \;\; E_{\mathrm{total}}
= \sum_{k=1}^{K} \gamma^{k-1} \, E_{\mathrm{total}}(k) \\
= \sum_{k=1}^{K} \gamma^{k-1} \Big(
    E_{\mathrm{movement}}(t_k)
  + E_{\mathrm{sensing}}(\mathtt{s}) \\
\quad
  + E_{\mathrm{computation}}(\alpha_{l_k}) \cdot (1 - o_{k}) \\
\quad
  + E_{\mathrm{transmission}}(k) \cdot o_{k}
\Big)
\end{split}
\end{equation}
\vspace{-10pt}
\begin{eqnarray} 
    \textrm{s.t.}& \eta_{\min} \leq \eta_{K}, \label{performance_const}
\end{eqnarray}
\vspace{-17pt}
\begin{flalign}
&& \mathtt{p}^{\min}_{k,i} < \mathtt{p}_{k,i} \le \mathtt{p}^{\max}_{k,i}, \;\forall \mathtt{p}_{k,i} \in \mathcal{P}_k,\; \forall i \in \{0,\dots,|\mathcal{P}_k|\},
\label{movement_range_const}
\end{flalign}
\vspace{-17pt}
\begin{eqnarray} &&
E_{\mathrm{total}}(k) \leq \beta_{k}. \label{energy_bounds}
\end{eqnarray}

The $\gamma \in [0,1]$ is the discount rate to account for the relative importance of the total energy consumption of future timesteps. $K$ denotes the terminal timestep at which the robot either successfully completes its mission or ends with false positive detection or battery depletion. Constraint (\ref{performance_const}) ensures a minimum acceptable performance of the selected AI algorithm at the terminal timestep. Constraint (\ref{movement_range_const}) ensures that the assigned motion command is bounded to robot's hardware specifications. Constraint (\ref{energy_bounds}) ensures that the total energy requirements for all operations at timestep $k$ do not exceed the available battery capacity of the robot.

To ensure constraints conformance, detection performance stability, and energy efficiency, a penalty function is introduced. It is comprised of two main terms, namely the $P_{\mathrm{constr}}(k)$, and $P_{\mathrm{det}}(k)$, account for constraints violation, and unsatisfactory detection performance, respectively, and applies to both the robot and the $\text{RC}$. The formulations are provided below:
\vspace{-5pt}



\vspace{-7pt}
\begin{equation}
\begin{aligned}
P_{\mathrm{constr}}(k) = \;&
    \sum_{i = 0}^{|\mathcal{P}_{k}|} 
   \max\Bigl(0, \; p_{k,i} - p^{\mathrm{max}}_{k,i}, \; p^{\mathrm{min}}_{k, i} - p_{k,i}\Bigr) \\
&+  \max\bigl(0, \; E_{\mathrm{total}}(k) - \beta_{k}\bigr),
\end{aligned}
\label{eq:constraint_penalty}
\end{equation}

\begin{equation}
    P_{\mathrm{det}}(k) = 1- \eta_{k}
    \label{eq:det_penalty}
\end{equation}

The total penalty is given by:
\begin{equation}
P_{\mathrm{total}}(k) = \lambda_1 \cdot P_{\mathrm{constr}}(k) + \lambda_2 \cdot P_{\mathrm{det}}(k)\;,
\label{eq:total_penalty}
\end{equation}
where $\lambda_1, \text{and} \; \lambda_2$ are the weights of the equation, controlling the impact of each penalty term.

Overall, the complete objective function that takes into account any introduced penalties is defined as:
\begin{equation}
 \min_{\mathcal{P}_{k}, o_{k}, l_{k}} \sum_{k=1}^{K} \gamma^{k-1} \cdot \big(\mu_1 \cdot E_{\mathrm{total}}(k) + \mu_2 \cdot P_{\mathrm{total}}(k)\big) 
\label{complete_obj_pen}
\end{equation}
where $\mu_1, \text{and} \; \mu_2$ are the weights of the equation, controlling the impact of each term. 
It should be noted that each term of the objective and penalty functions is min-max normalized and then divided by its mean to ensure comparable average contributions before applying the objective weights.

\section{ED$^3$R: Distributed Hierarchical $\epsilon$-Greedy Cooperative Agentic Intelligence}\label{proposed_algorithm}
This section introduces ED$^3$R, a distributed hierarchical $\epsilon$-greedy agentic decision-making to solve our optimization problem. The parameter $\epsilon \in [0,1]$ controls the probability between exploring and exploiting learned policies, where $\epsilon = 1$ corresponds to full exploration and $\epsilon = 0$ to full exploitation. The proposed approach consists of an offline warmup and an online phase. During both phases and at each timestep, the $\text{RC}$ and the robot observe different states of the environment and take different but interdependent actions that are jointly evaluated using  a reward function. 

\subsection{State, Action, Reward} \label{state_act_rew}
\textbf{State of the Environment:} Each agent has a different, yet complementary, view of the environment, defined as the state ($S$). The $\text{RC}$ maintains a higher-level perspective as it does not directly interact with the environment, whereas the robot operates with a concrete, interaction-driven view.

\underline{$\text{RC}$ State}: The $\text{RC}$, at each timestep, inspects a set of variables $S_{\mathrm{RC},k}$ that guides its decision-making. This set includes the robot’s position along the three axes ($x_{k-1}, y_{k-1}, z_{k-1}$) and pose, i.e. body's and sensors' orientation ($\mathtt{r}_{k-1} \; \text{and} \; \textbf{so}_{k-1}$), its distance from the robot ($\delta_{k-1,\text{RC}}$), as well as the SCS ($\eta_{{k-1}}$) and energy consumption ($E_{\mathrm{total}}({k-1})$) observed at the previous timestep, as an outcome of the preceding joint decisions of both agents. Overall:
    \[
    S_{\mathrm{RC},k} =
    \begin{aligned}
    \{\, &x_{k-1}, y_{k-1}, z_{k-1}, \mathtt{r}_{k-1}, \\&\textbf{so}_{k-1},\delta_{k-1, \text{RC}}, \eta_{{k-1}}, E_{\mathrm{total}}({k-1}) \,\}.
    \end{aligned}
    \]
        
\underline{Robot State}: The robot after executing RC's motion command, inspects its own set of parameters $S_k$ for its decision-making. These parameters include its pose ($\mathtt{r}_{k}, \textbf{so}_{k}$), the remaining battery ratio ($\beta^{\mathrm{ratio}}_{k}$), the cpu frequency ratio ($f^{\mathrm{ratio}}_{k}$), the distance from the $\text{RC}$ ($\delta_{k,\text{RC}}$), the allocated bandwidth ($b_{k}$), the achievable transmission data rate ($\mathtt{d}_{k}$), as well as the SCS ($\eta_{{k-1}}$) and energy consumption ($E_{\mathrm{total}}(k-1)$) observed at the previous timestep, as an outcome of the preceding joint decisions of both agents. Overall:
\begin{align}
    S_{k} =
    \{\, &\mathtt{r}_{k}, \textbf{so}_{k},\beta^{\mathrm{ratio}}_{k}, f^{\mathrm{ratio}}_{k}, \delta_{k,\text{RC}}, b_{k}, \mathtt{d}_{k}, \eta_{{k-1}}, E_{\mathrm{total}}(k-1)\,\}, \notag
\end{align}
    where
    \begin{equation}
        \beta^{\mathrm{ratio}}_{k} = \frac{\beta_{k}}{\beta^{\max}},
    \end{equation}
    \vspace{-7pt}
    \begin{equation}
        f^{\mathrm{ratio}}_{k} = \frac{f_{k}}{f^{\max}},
    \end{equation}
    and $f^{\max}$ denotes the maximum CPU frequency of the robot.

\textbf{Action Space}: The action space, is comprised of all decisions made by both agents at timestep $k$. 

\underline{$\text{RC}$ Actions}: As $A_{\mathrm{RC},k}$ we denote the action space of the $\text{RC}$, that represents assigned actions related to the movement of the robot along the three axes (${\delta^{(x)}_{k}, \delta^{(y)}_{k}, \delta^{(z)}_{k}}$), its body's and sensors' rotation ($\delta^{(\mathtt{r})}_{k}, \bm{\delta}^{\textbf{(so)}}_{k}$). Overall:
\begin{align}
A_{\mathrm{RC},k} = \{\delta^{(x)}_{k}, \delta^{(y)}_{k}, \delta^{(z)}_{k}, \delta^{(\mathtt{r})}_{k}, \bm{\delta}^{\textbf{(so)}}_{k}\}.\notag
\end{align}

\underline{Robot Actions}: As $A_{k}$ we denote the decision-making of the robot regarding the offloading of the detection task $o_{k}$ and the model selection $l_{k}$. 
Overall:
\vspace{-5pt}
\begin{align}
    A_{k} = \{o_{k}, l_{k}\}. \notag
\end{align}

\textbf{Mutual Reward}: The reward retrieved at the end of each $k$, mutual for both agents and based on Eq. (\ref{complete_obj_pen}), is defined as follows:
\vspace{-5pt}
\begin{equation}\label{reward}
    rr_k = - (\mu_1 \cdot E_{\mathrm{total}}(k) + \mu_2 \cdot P_{\mathrm{total}}(k)\big).
\end{equation}
Both agents cooperatively target to maximize the reward, which results on minimizing the total energy consumption with respect to the imposed detection confidence and the constraints of the environment.
\subsection{Offline Warmup Phase}\label{warmup_phase}
This phase enables both agents to acquire initial experience, and prevent a cold-start. The parameter $\epsilon$ is initialized to 1 for both agents, enforcing full exploration, and remains constant throughout the warmup phase. At each $k$, experience is acquired as follows:
\begin{enumerate}
    \item \textbf{Strategy Selection:} The $\text{RC}$ assigns a random motion command, encouraging broad exploration of the initially unknown environment. Upon execution, the robot independently selects random strategies (i.e., offloading versus local execution and model selection), ensuring full coverage of its action space.
    \item \textbf{Strategy Evaluation:} The actions of both agents are jointly evaluated using Eq.(\ref{reward}).
    \item \textbf{Data Logging:} All relevant information, including states, actions, and rewards, is recorded.
\end{enumerate}   
The warmup phase is repeated over a predefined number of missions until sufficient data is collected for agent training.

\textbf{Agent Forward-Looking and Training}: Observing the timeline within a timestep $k$ (Figure \ref{timeline_decisions}), both agents have to take sequential and complementary actions at different times. However, a single shared reward is retrieved only at the end of $k$, forcing each agent to \textit{look forward} by estimating the shared outcome of its actions before the complete decision sequence is executed. Most importantly, the $\text{RC}$, lacking knowledge of the robot’s subsequent actions, must \textit{look forward} to evaluate the outcome of its own decision while anticipating the expected effect of the robot’s subsequent decision on the final shared reward. This \textit{forward-looking} is the ability of both agents to anticipate the consequences of alternative actions by estimating outcomes that have not yet been observed at decision time \cite{Luo2024}.

To enable \textit{forward-looking}, one regression feedforward neural network (RFNN) model is trained for each agent in a supervised manner. Each model learns to estimate the resulting shared reward (model output) using as input features the agent’s respective state and actions, i.e. $S_{k} \cup A_{k}$ for the robot and $S_{\mathrm{RC},k} \cup A_{\mathrm{RC},k}$ for the $\text{RC}$. The rewards collected during this phase serve as ground truth for training. These trained models are used during the online phase for real-time execution.


\subsection{Online Real-Time Execution Phase}
During the online phase, the exploration parameter $\epsilon$ starts to decay by $\epsilon_{r}$ over time until reaching a minimum threshold $\epsilon_{\min}$, enabling a gradual transition from exploration to exploitation.

This phase leverages the experience obtained during the offline training to enable cooperative, real-time, chained decision-making between the two agents. The agents use their pre-trained RFNN models and follow a greedy-based strategy. At each $k$, the $\text{RC}$ evaluates a set of $N$ candidate alternative future actions\footnote{In large action spaces, exhaustive evaluation is infeasible, and as such a limited candidate set enables efficient local search.}, while the robot evaluates all possible actions. With probability $\epsilon$, each agent selects a random candidate action from the available ones, while with probability $1-\epsilon$, it exploits its acquired \textit{forward-looking} knowledge: it estimates the anticipated reward for each candidate action, and greedily selects the one that maximizes this reward.

To complement and accelerate cooperative decision-making during the online phase, ED$^3$R incorporates three additional mechanisms that promote safe operation while further enhancing exploration and energy efficiency. These mechanisms are detailed below.

\textbf{Obstacle Avoidance Mechanism (Robot):} The monitored environment is not obstacle-free. While the RC maintains a higher-level view of the environment, the robot has access to a more concrete local perspective and can detect nearby obstacles through onboard sensing, e.g., LiDAR. As such, it would be impractical for the robot to blindly follow every RC's motion command. ED$^3$R incorporates a 360$\degree$ obstacle avoidance mechanism at the robot side to ensure safe and reliable exploration of the surrounding environment. This mechanism suppresses motion commands that would drive the robot toward obstacles located within a minimum safety distance. Specifically, when an obstacle is detected along the direction of motion and its distance from the robot falls below the minimum safety threshold $\mathtt{sd}_{\mathrm{min}}$, the robot autonomously halts movement in this direction.

\textbf{Region Avoidance Mechanism (RC):} Since, the RC greedily selects among a number of candidate actions at each $k$, it may repeatedly select neighboring motion commands confining the robot's motion to a nearby vicinity. To enhance environment coverage efficiency, ED$^3$R introduces a region stuck avoidance mechanism at the RC side, driven by the SCS ($\eta_{k}$). The environment is partitioned into fixed-size cubic regions of size $\mathrm{crs}$. As the robot navigates across these regions, the RC maintains a region-wise logging history, which records the number of robot visits and the average SCS observed across visits. Based on this history, each region is categorized as: \textit{uninformative} (visited regions with zero average SCS), \textit{unexplored} (unvisited regions), and \textit{promising} (visited regions with non-zero average SCS). If the number of visits to an \textit{uninformative} region exceeds a predefined threshold $as_{\mathrm{thr}}$, we consider this state as local region stuck. 

Under local region stuck, we enforce region avoidance through a prioritization strategy. Priority is given to \textit{promising} regions based on logging history, by enforcing RC to greedily select an action that leads to the region with the maximum average SCS. If no \textit{promising} region is reachable through the available candidate actions, the RC selects an action that directs the robot toward an \textit{unexplored} region. Once the robot leaves the \textit{uninformative} region, standard ED$^3$R operation resumes.



\textbf{Adaptive Early Mission Completion Mechanism (RC)}: Even upon wildfire detection, the RC may select actions, which can cause abrupt perspective changes and unnecessary energy consumption. To counteract this, ED$^3$R employs an adaptive early mission completion mechanism, improving energy efficiency through motion stabilization or mission termination, when the SCS indicates that further motion is unnecessary. Specifically, under reliable detections (line \ref{reliability_check}), the RC commands the robot to hover in place while continuing its perception tasks. If this condition persists for $\mathrm{esp}$ consecutive timesteps, the mission completes successfully (line \ref{termination_criterion}). For intermediate confidence levels (line \ref{intermediate_reliability}), and to prevent abrupt changes in robot's perspective, the RC's action value ranges, denoted by $\mathcal{A}_{\mathrm{RC},k}$, are decreased by a fixed value $\text{fv}_a \in \textbf{fv}, \forall a \in \mathcal{A}_{\mathrm{RC},k}$ (line \ref{decrease_factor}). If intermediate confidence persists for $c_{\mathrm{thr}}$ consecutive timesteps, more drastic corrective measures are applied: the action set is expanded using the same value, and exploration is enabled through $\epsilon$ (line \ref{increase_factor}). In case the SCS falls below the intermediate confidence threshold, the action value ranges are reset and standard ED$^3$R operation resumes (line \ref{dsar_resume_actions}).


\begin{algorithm}[!ht]
\small
\captionsetup{labelformat=empty}
\caption{\textbf{Algorithm}: ED$^3$R Adaptive Early Mission Completion}
\label{alg:early_stopping}
\begin{algorithmic}[1]
\Require $\eta_{\min}, \eta_{\mathrm{int}}, \mathrm{esp}, c_{\mathrm{thr}}, \textbf{fv}$
\State Initialize $c_{\mathrm{stop}}\gets0$, $c_{\mathrm{low}}\gets0$
\For{each timestep $k$}
    \If{$\eta_{k-1}\geq\eta_{\min}$} \label{reliability_check}
        \State \textit{hover}; $c_{\mathrm{stop}}\gets c_{\mathrm{stop}}+1$; $c_{\mathrm{low}}\gets0$; $\epsilon\gets0$ 
        \State \textbf{if} $c_{\mathrm{stop}}\geq \mathrm{esp}$ \textbf{then} \textit{complete mission} \label{termination_criterion}
    \ElsIf{$\eta_{\mathrm{int}}\leq\eta_{k-1}<\eta_{\min}$} \label{intermediate_reliability}
        \State $c_{\mathrm{stop}}\gets0$; $c_{\mathrm{low}}\gets c_{\mathrm{low}}+1$
        \If{$c_{\mathrm{low}} < c_{\mathrm{thr}}$} 
             \State $\mathcal{A}_{\mathrm{RC},k}\gets\{a/\mathrm{fv}_a\mid \text{fv}_a \in \textbf{fv}, a\in \mathcal{A}_{\mathrm{RC},k}\}$; $\epsilon\gets0$ \label{decrease_factor}
        \Else \label{intermediate_persists}
            \State $\mathcal{A}_{\mathrm{RC},k}\gets\{a + \mathrm{fv}_a\mid \text{fv}_a \in \textbf{fv},  a\in \mathcal{A}_{\mathrm{RC},k}\}$; $\epsilon\gets1$ \label{increase_factor}
        \EndIf

    \Else
        \State \textit{reset} $\mathcal{A}_{\mathrm{RC},k}$; $c_{\mathrm{stop}}\gets0$; $c_{\mathrm{low}}\gets0$; $\epsilon\gets0$ \label{dsar_resume_actions}
    \EndIf
\EndFor
\end{algorithmic}
\end{algorithm}

\textbf{Extension to Multi-Robot Cooperation}:
Although the current system considers a single UAV robot and an RC, the distributed approach of ED$^3$R can be extended to a multi-robot setup. In such a setup, each robot could make its own decisions based on local observations (state), while the coordination of their motion could be handled by one or multiple distributed RC agents that maintain an aggregated view of the monitored area(s). The forward-looking mechanism could accordingly be extended so that robot agents evaluate the anticipated outcome of their local decisions. The RC agent(s) would evaluate strategies by accounting for the expected subsequent decisions of multiple robots and their contribution to the shared mission objective. Such an extension would require additional mechanisms for inter-robot coordination and cooperation, including information sharing, and joint exploration. 

\section{Performance Evaluation} \label{performance_evaluation}

\textbf{Mission}: A single UAV robot is equipped with a set of sensors for wildfire detection and safe navigation, i.e. optical camera, LiDAR, GPS, Inertial Measurement Unit (IMU), and two computer vision (CV) detection models. The robot cooperates with one RC, having its own CV model, capable of controlling the robot's motion along axes, rotation, and camera roll. Together, these two agents are tasked to detect a wildfire incident with minimum energy consumption, while satisfying an imposed detection confidence threshold $\eta_{\mathrm{min}}$. 

\textbf{Aim}: Prove that energy-optimal autonomy under uncertainty requires cooperative, hierarchically chained decision-making across control, perception, computation, communication, and energy management. Accordingly, we structure the evaluation around three key points: \textbf{1)} the energy-mission reliability trade-off, \textbf{2)} the importance of hierarchical coordination and \textbf{3)} the effectiveness of ED$^3$R compared to centralized and heuristic baselines. 

\textbf{Evaluation Metrics}: To quantitatively evaluate the performance of ED$^3$R, Table \ref{tab:acronyms} summarizes the considered metrics.
The MSR expresses the percentage of successfully completed missions, excluding battery depletions (BDs) or False Positives (FPs). The Wasted Energy Rate (WER) quantifies the percentage of wasted energy relative to the accumulated total energy consumption. The look-forward (LF) accuracy (\%) is expressed as:
\vspace{-5pt}
\begin{equation}
    \text{LF Accuracy} = (1- \frac{MAE_{rr}}{MAE^{\mathrm{max}}_{rr}}) \cdot 100\%, \notag
\end{equation}
where the $MAE_{rr}$ expresses the mean absolute error between the anticipated and actual reward in a single experiment, and $MAE^{\mathrm{max}}_{rr}$ expresses the maximum MAE across experiments. 

\begin{table}[!ht]
\centering
\resizebox{\columnwidth}{!}{%
\begin{tabular}{|
>{\columncolor[HTML]{EFEFEF}}l |l|}
\hline
\multicolumn{1}{|c|}{\cellcolor[HTML]{EFEFEF}\textbf{Metric}} & \multicolumn{1}{c|}{\cellcolor[HTML]{EFEFEF}\textbf{Definition}} \\ \hline
\textbf{Total Energy (kJ)} & Total Energy consumed during robot operation \\ \hline
\textbf{Computation Energy (kJ)} & Energy consumed during CV execution \\ \hline
\textbf{Communication Energy (kJ)} & Energy consumed during sensing data transmission \\ \hline
\textbf{Movement Energy (kJ)} & Energy consumed during robot movement \\ \hline
\textbf{Sensing Energy (kJ)} & Energy consumed during sensing \\ \hline
\cellcolor[HTML]{EFEFEF}\textbf{MSCS (\%)} & Mean Stable Confidence Score over a mission \\ \hline
\textbf{MSR (\%)} & Mission Success Rate \\ \hline
\textbf{LF Accuracy (\%)} & Look-Forward Accuracy \\ \hline
\cellcolor[HTML]{EFEFEF}\textbf{WER (\%)} & Wasted Energy Rate \\ \hline
\textbf{OR (\%)} & Offloading Rate \\ \hline
\cellcolor[HTML]{EFEFEF}\textbf{FPs (\%)} & False Positives rate \\ \hline
\cellcolor[HTML]{EFEFEF}\textbf{BDs (\%)} & Battery Depletion rate \\ \hline
\cellcolor[HTML]{EFEFEF}\textbf{Mission Duration (minutes)} & Total flight time until mission completion \\ \hline
\end{tabular}%
}
\caption{Evaluation Metrics Definition}
\label{tab:acronyms}
\vspace{-5pt}
\end{table}

\subsection{Simulation Setup}\label{simulation_setup}

This section outlines our simulation ecosystem, including: \textbf{1)} the 3D environment, \textbf{2)} the UAV robot, \textbf{3)} the CV detection models, and \textbf{4)} the wireless communication network.

\begin{figure}[!ht]
    \centering
    \includegraphics[width=0.7\linewidth]{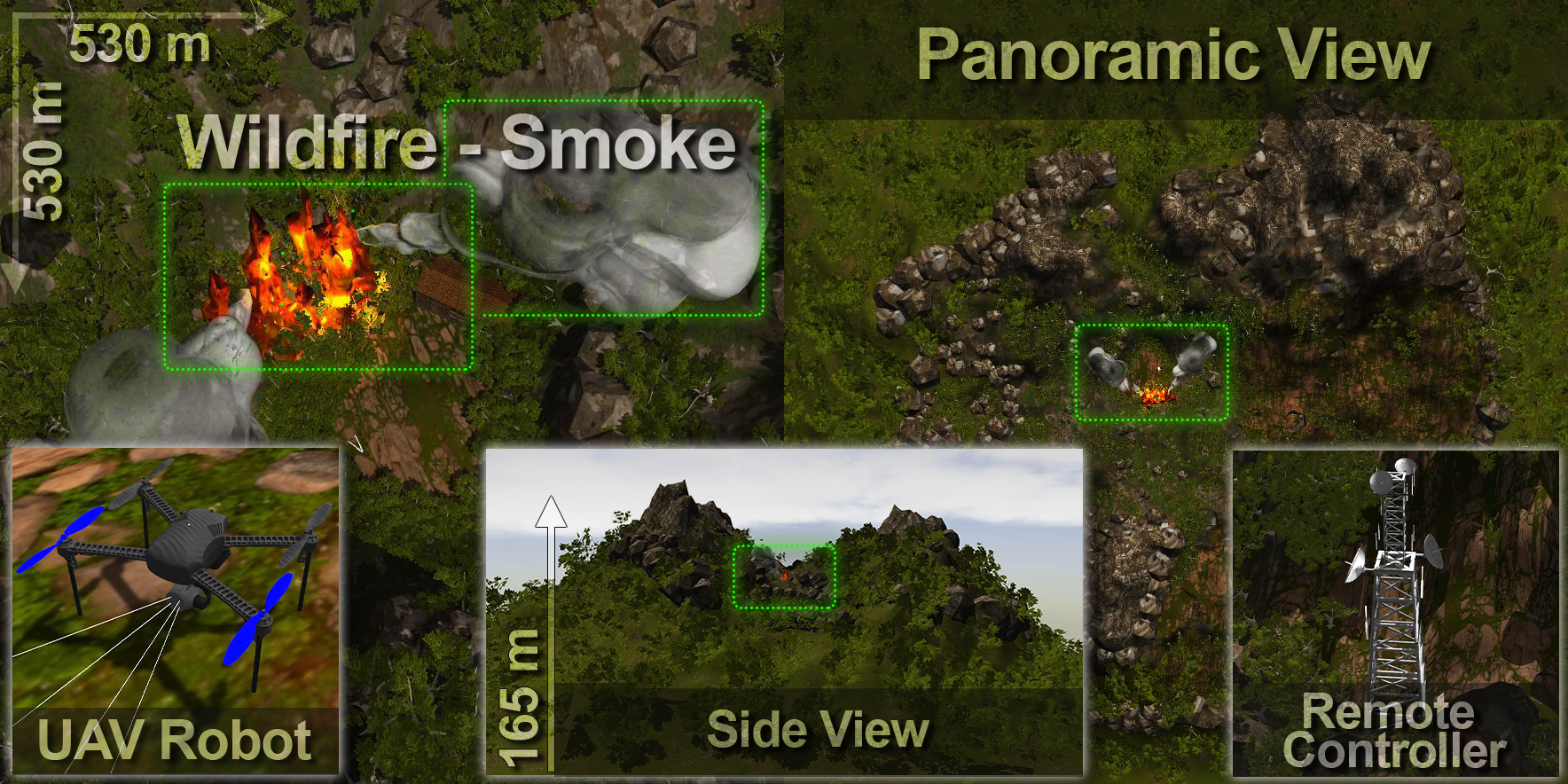}
    \caption{Simulation Environment inside Gazebo}
    \label{gazebo_environment}
\end{figure}

\textbf{3D Environment}: The simulation environment is built upon Gazebo 11 \cite{gazebo} and Robot Operating System 2 (ROS2) \cite{ros2}, incorporating rigid-body dynamics and collision physics through the Open Dynamics Engine \cite{ode}. These components provide a physics-aware simulation setting for evaluating robotic behavior under realistic terrain and environmental conditions. Figure \ref{gazebo_environment} illustrates the environment that includes a custom 530 x 530 x 165 m (width x depth x height) mountainous forest scene along with a 46 m high wildfire (including smoke), all designed in Blender \cite{blender}.

\textbf{UAV Robot}: The simulated UAV properties were highly inspired by the 3DR Iris quadcopter to provide a realistic reference for the robot model\cite{iris_drone}. Table \ref{tab:robot_setup} given in the Appendix, summarizes the UAV robot's configuration, including its set of sensors, hardware specifications, computational and communication capabilities, and key physical properties.

\textbf{CV Detection Models}: Two custom convolutional neural network (CNN) models are available for wildfire detection: a Full CNN and a Quantized CNN, differing in computational complexity. Both models are available at the UAV robot, while the $\text{RC}$ uses only the Full one. Details regarding their training and configuration are provided in the Appendix (Table \ref{tab:cv_models}).

\textbf{Wireless Communication Network}: The communication model considers a single Base Station (BS). The $\text{RC}$ is assumed to be collocated with the BS, and as such share the same physical location and jointly represent the communication endpoint of the UAV robot. A 2.4 GHz wireless link connecting the robot and the RC is characterized by distance-dependent path loss, shadow fading, bounded bandwidth, and spectral efficiency. The transmit power is determined with respect to a target received signal strength, while the complete network parameterization is provided in Appendix (Table \ref{tab:network_setup}).

\textbf{Ecosystem Dynamicity}:  
To simulate realistic conditions, introduce dynamicity and uncertainty into the environment and evaluate the generalization ability of ED$^3$R, several key parameters are varied at each timestep. These variations capture changes in resource availability at the robot (CPU speed, assigned bandwidth), as well as in wireless communication quality, including degradation with distance and fluctuations in the achievable data rate and transmission power requirements (Section \ref{problem_formulation}). The environment includes 14 distinct wildfire locations, with one wildfire incident per experiment, and two possible BS locations. The distance between each wildfire location and the corresponding BS ranges from 18 to 275 m. The wildfire locations introduce detection difficulty (e.g. distant and obscured), while alternative BS placements impact connectivity and detection reliability.

\textbf{Simulation Realism - Deployment Challenges}:
Although the evaluation is simulation-based, the considered ecosystem captures several key aspects of robotic deployment, including UAV dynamics, realistic energy consumption models, heterogeneous sensing and computation capabilities, and time-varying network conditions. However, real-world deployment introduces additional challenges that cannot be fully reproduced in simulation, such as sensor noise, unexpected communication interruptions, hardware limitations, and inconsistencies between modeled and actual energy consumption. These factors may affect detection confidence and decision-making reliability. Future work will investigate the deployment of ED$^3$R on physical robotic hardware and evaluate its performance under real-world operational conditions.

\subsection{ED$^3$R Configuration Setup}\label{simulation_setup}

Table \ref{tab:dsar_setup} summarizes the configuration of ED$^3$R regarding the region size, motion command ranges, thresholds and key algorithmic parameters. Table \ref{tab:DSAR-NN-Configuration} summarizes the configuration for the RFNN models used by the UAV Robot and the $\text{RC}$. The RFNN training samples were generated according to the offline warmup described in Section \ref{warmup_phase}. The wildfire location was sampled uniformly for each experiment, while agent decisions were sampled uniformly at each timestep. The robot’s available computational and communication resources were uniformly resampled every 5 timesteps within their configured ranges (Table \ref{tab:robot_compute}). Different mission durations resulted in unequal samples per experiment, ranging from 5 to 200. Overall, 115K samples, a batch size of 128, the Adam optimizer with learning rate 5e-06 and 2000 epochs were used for the offline training phase with 85\%, 10\%, and 5\% of the samples used for the training, validation and evaluation. To avoid overfitting, dropout layers were used with 10\% probability of zeroing neural connections along with an early stopping where the patience was set to 10 epochs. All hyperparameters were empirically tuned during the offline warmup phase, based on mission success, energy consumption, detection reliability, and training stability. The selected values were then fixed for all online evaluation experiments.

\begin{table}[!ht]
\centering
\resizebox{220pt}{!}{%
\begin{tabular}{|
>{\columncolor[HTML]{EFEFEF}}l |l|l|}
\hline
\textbf{Entity RFNN} & \cellcolor[HTML]{EFEFEF}\textbf{\# Input Features} & \cellcolor[HTML]{EFEFEF}\textbf{\# RFNN Layers: {[}Shape{]}} \\ \hline
\textbf{UAV Robot} & 11 ($S_{k} \cup A_{k}$) & 3: {[}128 x 64 x 32{]} \\ \hline
\textbf{$\text{RC}$} & 13 ($S_{\mathrm{RC},k} \cup A_{\mathrm{RC},k}$) & 4: {[}128 x 64 x 64 x 32{]} \\ \hline
\end{tabular}%
}
\caption{ED$^3$R: RFNNs for the UAV Robot \& $\text{RC}$}
\label{tab:DSAR-NN-Configuration}
\end{table}

\begin{table}[!ht]
\centering
\resizebox{220pt}{!}{%
\begin{tabular}{|
>{\columncolor[HTML]{EFEFEF}}l |l|}
\hline
\multicolumn{1}{|c|}{\cellcolor[HTML]{EFEFEF}\textbf{Parameter}} & \multicolumn{1}{c|}{\cellcolor[HTML]{EFEFEF}\textbf{Value}} \\ \hline
\textbf{Cubic Region Size} ($\mathrm{crs}$) & 60x60x60 m \\ \hline
\textbf{Axes Movement Range}($\delta^{(x)},\delta^{(y)}, \delta^{(z)}$) & [-20,20] m \\ \hline
\textbf{Body Rotation Range} ($\delta^{(\mathtt{y})}$)& [-90,90]$\degree$  \\ \hline
\textbf{Camera Roll Range} ($\bm{\delta}^{\textbf{(so)}}$)& [0,90]$\degree$  \\ 
\hline
\textbf{\# Candidate Actions} ($N$) & 1000 \\ \hline
\textbf{Penalty Weights} ($\lambda_1,\lambda_2$) & 1,1 \\ \hline
\textbf{Objective Weights} ($\mu_1,\mu_2$) & 1,2 \\ \hline
\textbf{Intermediate Confidence Threshold} ($\eta_{int}$) & 20\% \\ \hline
\textbf{Decrease Factors} (\textbf{fv}) & [3,10,10] \% \\ \hline
\textbf{Epsilon min} ($\epsilon_{\mathrm{min}}$) & 0.01 \\ \hline
$\epsilon$-\textbf{Decay Rate }($\epsilon_{r}$) & 0.95 \\ \hline
\textbf{Region Stuck Avoidance Threshold} ($as_{\mathrm{thr}}$) & 20 \\ \hline
\textbf{Early Mission Completion Patience} ($\mathrm{esp}$) & 2 \\ \hline
\textbf{Consecutive Low Confidence Threshold} ($c_{\mathrm{thr}}$) & 2 \\ \hline
\end{tabular}%
}
\caption{ED$^3$R: Configuration Setup}
\label{tab:dsar_setup}
\vspace{-10pt}
\end{table}

\subsection{Evaluation Results}
The evaluation is conducted through a mission-efficiency perspective. During the offline warmup phase, only 6 out of the 14 wildfire locations and a single BS location are used. As such, during the online real-time execution phase, ED$^3$R is evaluated under known and unseen event and BS placements and communication conditions, thereby assessing its generalization capability. During the warmup phase 2040 missions/experiments are conducted  and 850 during the online phase. For clarity, only 640 warmup experiments are presented.

\textbf{Result 1 - Adaptive Perception Efficiency}: This first round of results tests the performance of ED$^3$R under varying detection performance thresholds ($\eta_{\mathrm{min}}$): 85\% (high perception), 75\% (medium perception), 65\% (low perception).

\begin{figure}[!ht]
    \centering
    \includegraphics[width=0.93\linewidth]{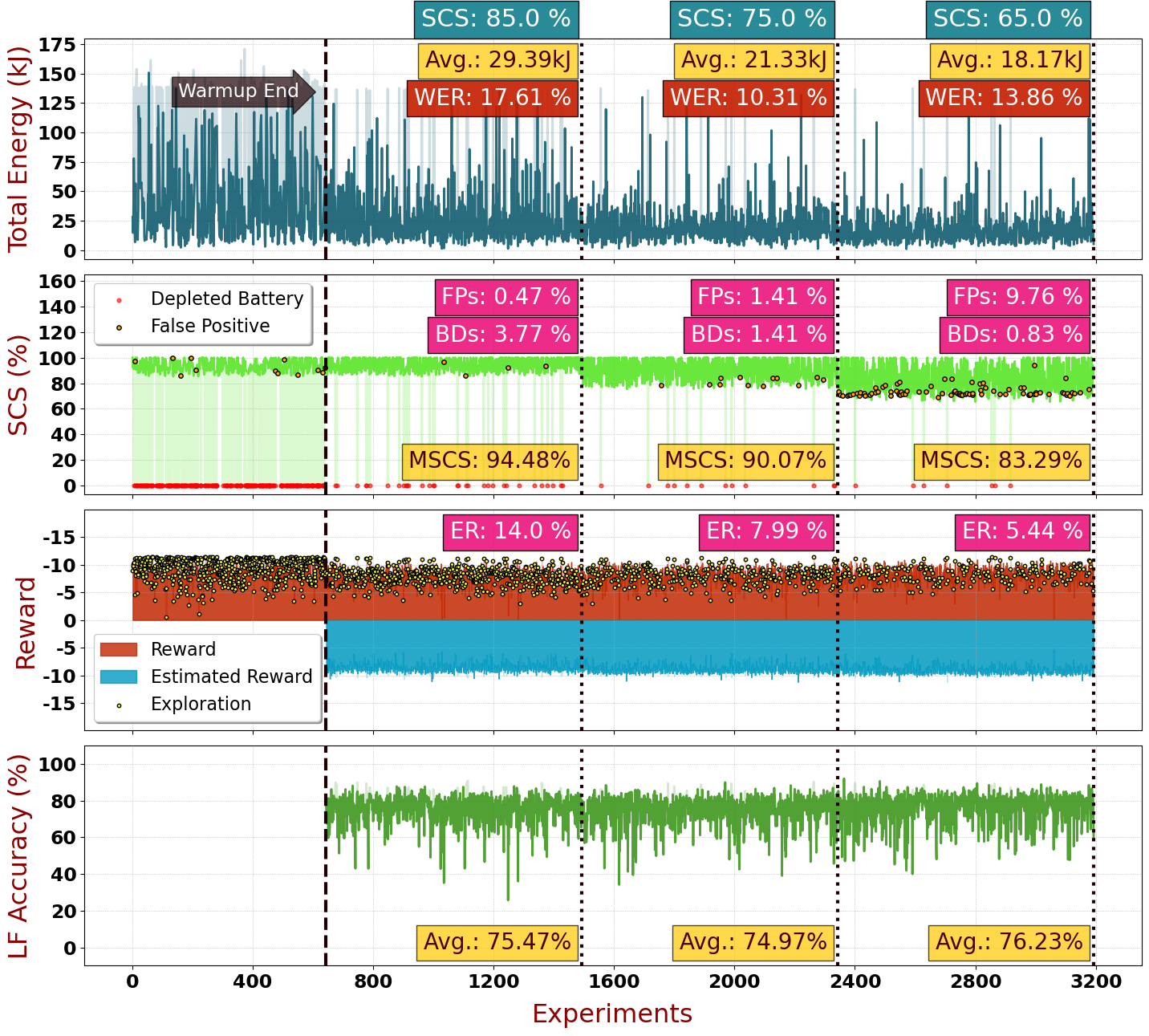}
    \caption{Performance overview of ED$^3$R under varying detection performance thresholds}
    \label{fig:reward_performance1}
    \vspace{-7pt}
\end{figure}

Figure \ref{fig:reward_performance1} presents an overview of the ED$^3$R evaluation in terms of detection reliability, energy consumption and \textit{forward-looking} capability. Each point in the figures represents an experiment. The dashed line marks the transition from the warmup phase to the online phase, while the dotted lines denote the boundaries between different performance thresholds, as indicated at the top of the figure (blue boxes). The statistics have been computed over a total of 850 experiments per detection performance threshold, each of which was conducted independently. The low opacity lines indicate statistics of battery depletions.

\noindent\textbf{Figure \ref{fig:reward_performance1} - Remarks:}
\begin{itemize}
    \item Results across confidence thresholds reveal a clear trade-off between energy efficiency and detection reliability. Higher thresholds (85\%) increase average energy consumption (29.39 kJ), battery depletions (BDs: 3.77\%), and exploration rate (ER: 14\%), reflecting more conservative behavior, but yield the lowest false positive rate (FPs: 0.47\%) and the highest detection confidence (MSCS: 94.48\%). In contrast, lower thresholds (65\%) reduce energy consumption (18.17 kJ, $\downarrow$ 38.2\%), battery depletions, and exploration (ER: 5.44\%), but at the cost of higher false positives (FPs: 9.76\%) and lower detection confidence (MSCS: 83.29\%, $\downarrow$ 12\%).
    \item The intermediate threshold (75\%) provides the most balanced performance, achieving an average LF accuracy of 75.47\%, a substantial reduction in energy consumption (21.33 kJ, $\downarrow$ 27.4\%) and wasted energy (WER: 10.31\%) while maintaining low false positives (1.41\%) and high detection reliability (MSCS: 90.07\%). This balance is also reflected in the limited number of battery depletions (BDs: 1.41\%), indicating that moderate confidence thresholds enable efficient yet reliable operation.

\end{itemize}

\begin{figure}[!ht]
    \centering
    \includegraphics[width=0.93\linewidth]{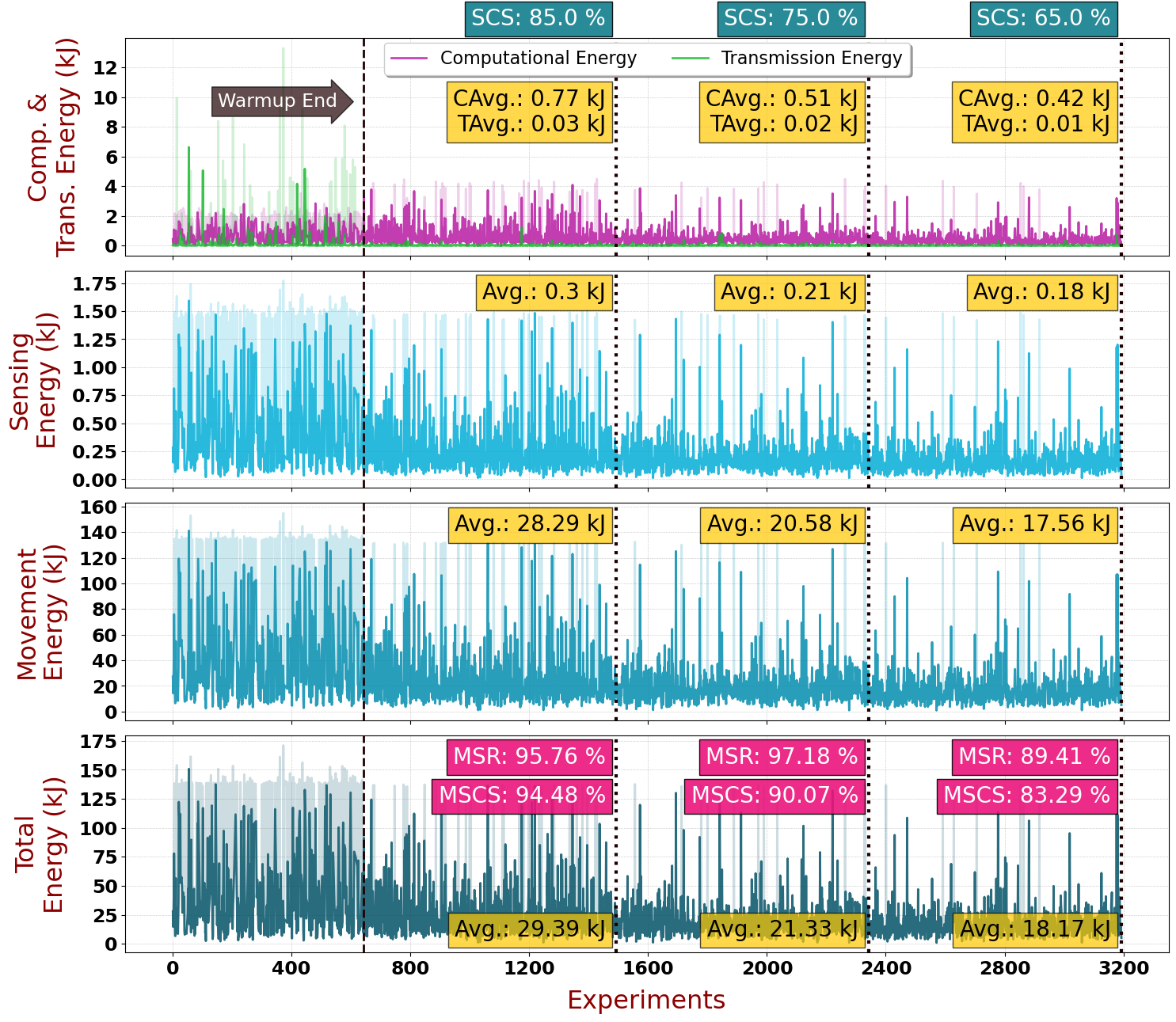}
    \caption{Performance Insights of ED$^3$R under varying detection performance thresholds}
    \label{fig:reward_performance2}
    \vspace{-7pt}
\end{figure}

Figure \ref{fig:reward_performance2} complements the previous analysis by decomposing the total energy consumption into its main components and relating them to mission outcomes.

\noindent\textbf{Figure \ref{fig:reward_performance2} - Remarks:}
\begin{itemize}
    \item Movement energy remains the dominant contributor across all thresholds (28.29, 20.58, 17.56 kJ). However, this is not solely due to movement commands. It is also indirectly influenced by UAV-side decisions, as choosing between local execution and offloading can change hovering duration and, in turn, increase or reduce movement-related energy consumption.
    \item Computation (Cavg), transmission (Tavg), and sensing energy remain comparatively small ($\sim$3\% of the total energy), but their effect is still important as they not only influence the immediate reward but also the hovering time (more details on this aspect under \textbf{Result 2}). 
    \item In terms of mission success rate, the 75\% threshold achieves the highest MSR (97.18\%), while the 65\% threshold results in a clear drop (MSR: 89.41\%).
\end{itemize}

\begin{figure}[!ht]
    \centering
    \includegraphics[width=0.93\linewidth]{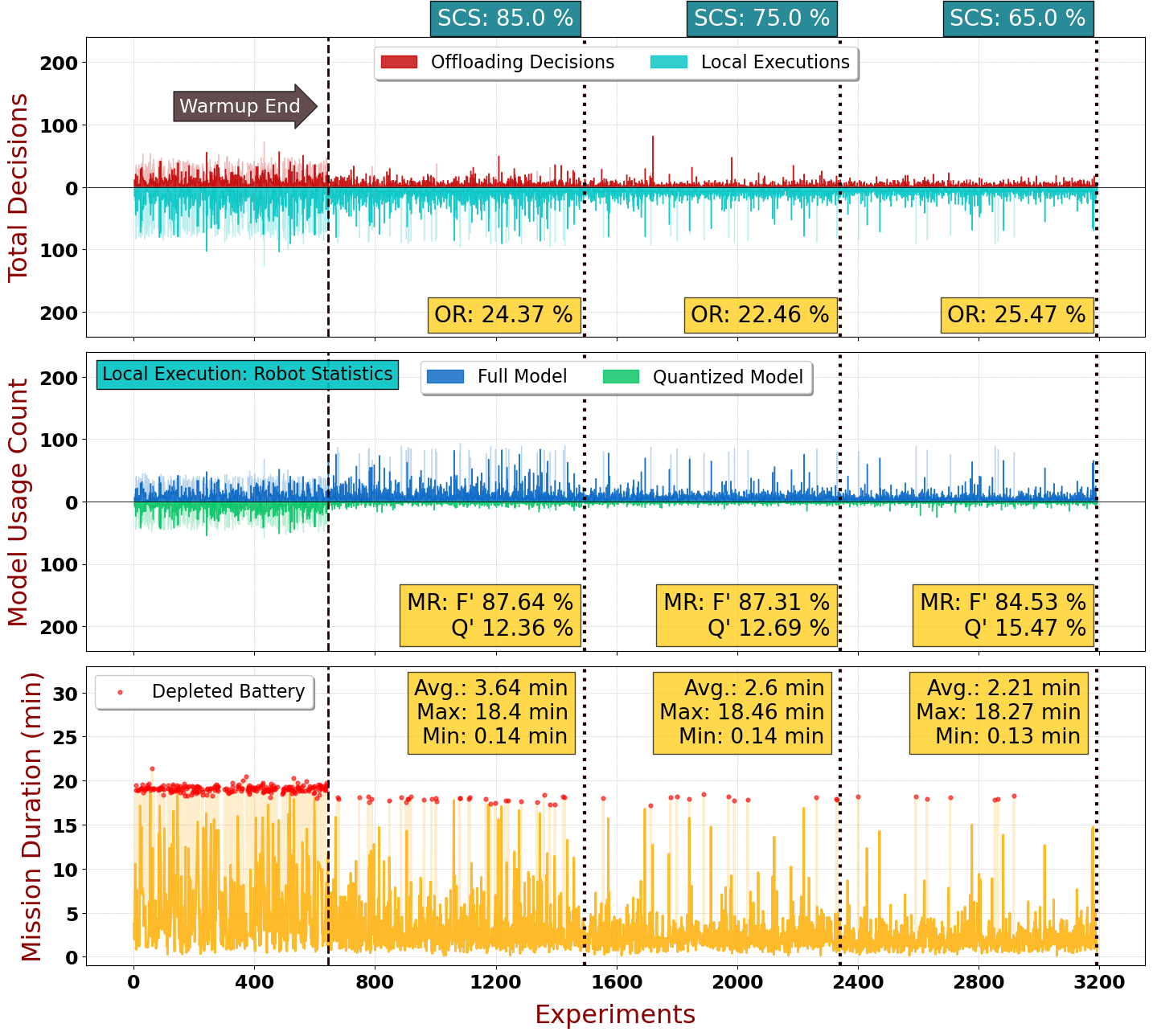}
    \caption{UAV robot's decision-making under varying detection performance thresholds}
    \label{fig:reward_performance3}
    \vspace{-8pt}
\end{figure}

Figure \ref{fig:reward_performance3} further explains Figures \ref{fig:reward_performance1} and \ref{fig:reward_performance2} by reporting the UAV robot-side decisions of ED$^3$R.

\noindent\textbf{Figure \ref{fig:reward_performance3} - Remarks:}
\begin{itemize}
    \item The offloading rate remains relatively stable across thresholds (OR: 24.37\%, 22.46\%, and 25.47\%), indicating that ED$^3$R does not rely heavily on offloading. Instead, it follows a selective offloading strategy, using communication only when beneficial under current environmental conditions.
    \item Across all thresholds, the model selection rate (MR) shows that Full model (F) is selected most often (87.64\%$\vert$12.36\%, 87.31\%$\vert$12.69\%, and 84.53\%$\vert$15.47\% for Full$\vert$Quantized), indicating that ED$^3$R generally prioritizes higher detection performance. At the same time, usage of Quantized model (Q) increases at the 65\% threshold, suggesting that lower thresholds favor energy efficiency over performance in absence of strict performance requirements.
    \item Average mission duration decreases from 3.64 to 2.21 minutes as threshold is relaxed, indicating that lower thresholds enable earlier mission completion. In contrast, the 85\% threshold remains conservative, spending more flight time verifying detections.
    \item The maximum mission duration remains at $\sim$18 minutes across all thresholds, as hard cases still require longer exploration. However, the 10-minute critical detection window (Introduction, EUSPA) is satisfied by a high percentage of missions, reaching the $85^{th}$, $95^{th}$, and $97^{th}$ percentiles as percentage threshold relaxes. Also, the maximum flight times are consistent with the endurance range of small UAVs, supporting the realism of the system model, problem formulation, and simulation configuration without assuming idealized flight conditions. 
\end{itemize}
\textbf{Conclusion 1}: ED$^3$R remains effective and realistic across all detection performance thresholds, consistently achieving high mission success while adaptively balancing energy use with performance requirements.

The 75\% confidence threshold strikes a balance between energy consumption, detection performance, mission duration and success rate. Therefore, for all subsequent results, ED$^3$R will be configured using this threshold.

\begin{figure}[!ht]
    \centering
    \includegraphics[width=0.7\linewidth]{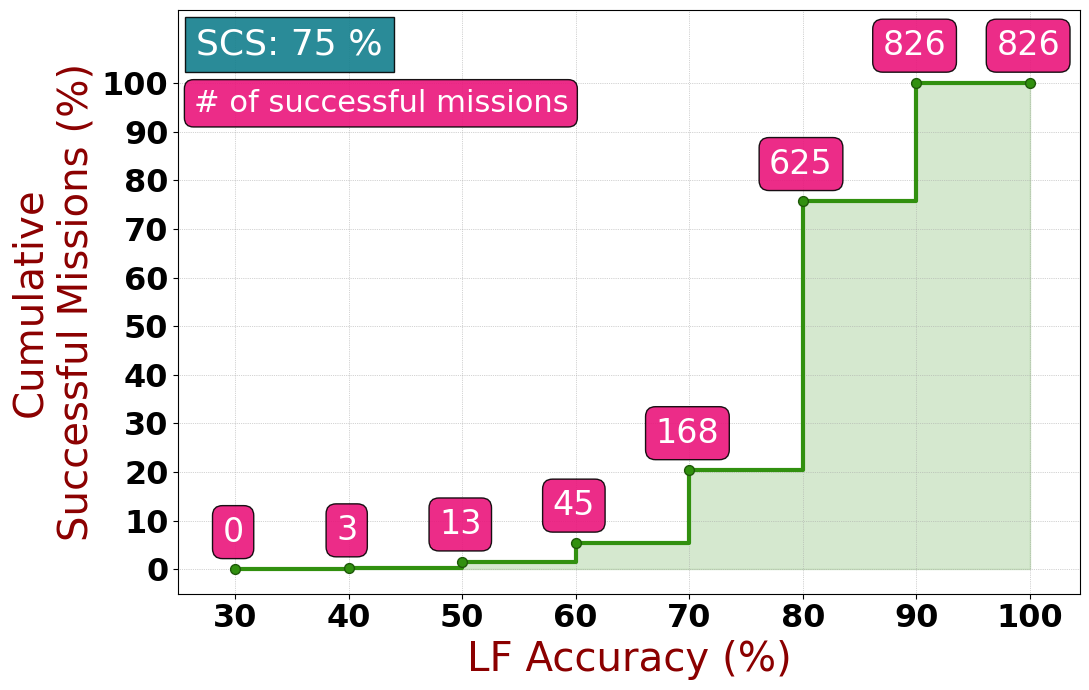}
    \caption{Correlation between LF Accuracy \& Cumulative MSR}
    \label{fig:correlation}
\end{figure}

\textbf{Result 2 - Impact of Forward-Looking}:
This second round of results investigates the importance of ED$^3$R's forward-looking capability. In essence, the LF accuracy measures how accurately the agents anticipate the impact of future actions, whether this impact is favorable or not for the mission. To capture this impact on final mission outcomes, Figure \ref{fig:correlation} illustrates the cumulative distribution of successful missions across LF accuracy levels. A clear positive relationship can be observed. As LF accuracy increases, successful missions accumulate more rapidly, with the highest concentration occurring above 70\%  accounting for 658 out of 826 successful missions).\\
\textbf{Conclusion 2}: More accurate predictions are not only associated with more reliable agent decision-making, but also with higher mission success.

\begin{figure}[!ht]
    \centering
    \includegraphics[width=1\linewidth]{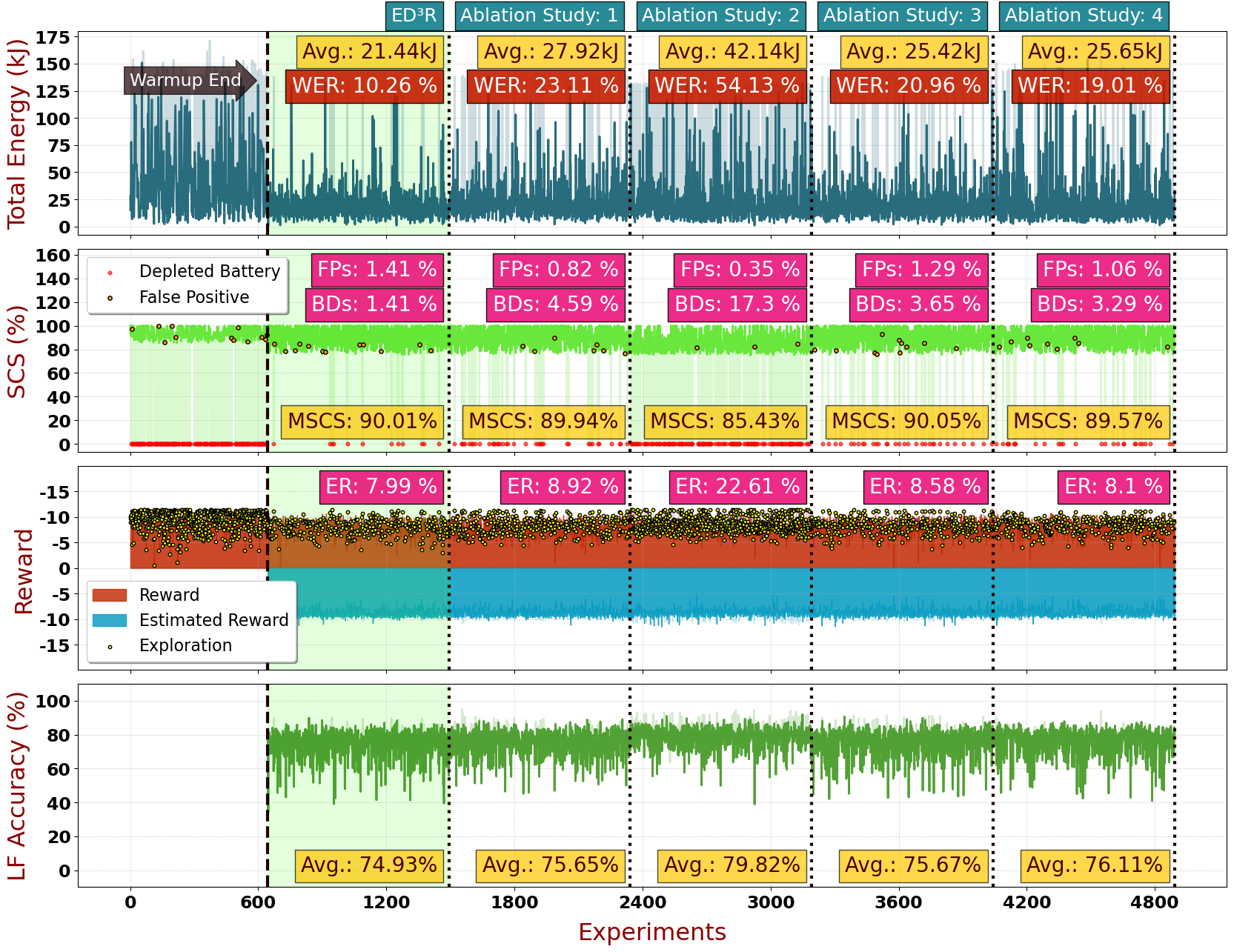}
    \caption{Performance overview under 4 ablation studies}
    \label{fig:abl_reward_performance1}
\end{figure}

\begin{figure}[!ht]
    \centering
    \includegraphics[width=1\linewidth]{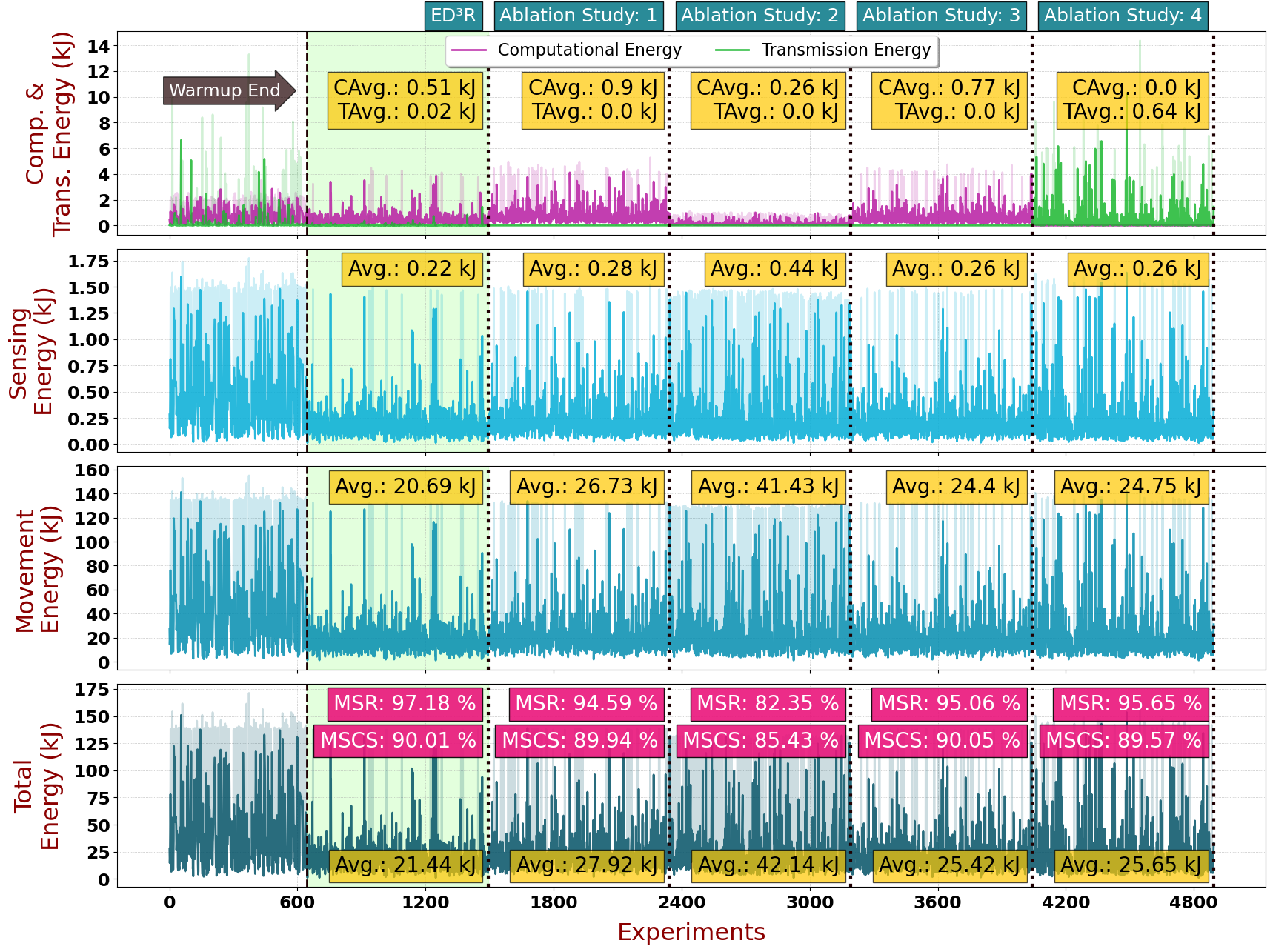}
    \caption{Performance insights under 4 ablation studies}
    \label{fig:abl_reward_performance2}
\end{figure}

\textbf{Result 3 - Ablation Studies - The importance of Cooperative Decision-Making}: During this third round of results, ED$^3$R's online mechanisms remain enabled as their joint operation supports safe and effective mission execution. More specifically, obstacle and region avoidance improve safety and exploration and can produce more informative observations. Informative observations can then enable early mission completion, and subsequently early completion contributes to the overall energy reduction through motion stabilization or mission termination. As such, this evaluation round focuses on demonstrating the joint effect of distributed decision-making and cooperation. Four ablation studies (Table \ref{tab:ablation}) are conducted, systematically disabling key components to reveal their individual impact and quantify their contribution to the overall performance. Together, these ablations cover the full set of individual decision-making capabilities across both agents, allowing us to evaluate each agent both independently and within the cooperative ecosystem. Figures \ref{fig:abl_reward_performance1} and \ref{fig:abl_reward_performance2} illustrate the outcome of the comparison analysis.

\begin{table}[!ht]
\centering
\resizebox{220pt}{!}{%
\begin{tabular}{|
>{\columncolor[HTML]{EFEFEF}}l |p{0.7\columnwidth}|}
\hline
\multicolumn{1}{|c|}{\cellcolor[HTML]{EFEFEF}\textbf{Ablation Study}} & \multicolumn{1}{c|}{\cellcolor[HTML]{EFEFEF}\textbf{Description}} \\ \hline
\textbf{Ablation Study 1} & \textbf{$\text{RC}$ agent}: $\text{RC}$'s RFNN model for motion command. \textbf{Fixed robot decision}: always local execution using the Full CNN model. \\ \hline
\textbf{Ablation Study 2} & \textbf{$\text{RC}$ agent}: $\text{RC}$'s RFNN model for motion command. \textbf{Fixed robot decision}: always local execution using the Quantized CNN model. \\ \hline
\textbf{Ablation Study 3} & \textbf{$\text{RC}$ agent}: $\text{RC}$'s RFNN model for motion command. \textbf{Robot agent}: always local execution, selecting from all available CNN models. \\ \hline
\textbf{Ablation Study 4} & \textbf{$\text{RC}$ agent}: $\text{RC}$'s RFNN model for motion command. \textbf{Fixed robot decision}: always offloading to $\text{RC}$. \\ \hline
\end{tabular}
}
\caption{Ablation Studies}
\label{tab:ablation}
\end{table}

\noindent\textbf{Figures \ref{fig:abl_reward_performance1} \& \ref{fig:abl_reward_performance2} - Remarks:}
\begin{itemize}
\item ED$^3$R achieves the best overall trade-off, with the lowest total energy consumption (21.44 kJ), lowest WER (10.26\%), lowest BDs and FPs (1.41\%), and highest MSR (97.18\%). These results show that RC-robot cooperation is key to maintaining reliable detection with low energy cost.
\item Ablation Study 1 shows that relying on the highest-complexity local model preserves detection quality, with MSCS close to ED$^3$R, but at a clear energy cost. Total energy increases to 27.92 kJ, mainly due to longer computation and hovering time, while WER and BDs rise to 23.11\% and 4.59\%, respectively.
\item Ablation Studies 2-4 show that fixed execution strategies, whether local or fully offloaded, are insufficient. Ablation Study 2 performs worst, reaching the highest energy consumption (42.14 kJ), WER (54.13\%), and BDs (17.3\%). Although Ablation Studies 3 and 4 reduce energy compared to Ablation Study 2, both still nearly double the WER of ED$^3$R, confirming the need for adaptive robot-side execution.

\item ED$^3$R obtains a slightly lower LF accuracy (74.93\%) than the ablations due to adaptive robot decisions making them harder to predict. In contrast, the ablations follow fixed decisions that make reward estimation easier, yet this higher predictability does not translate into better mission outcomes.
\end{itemize}

\begin{figure}[!ht]
    \centering
    \includegraphics[width=1\linewidth]{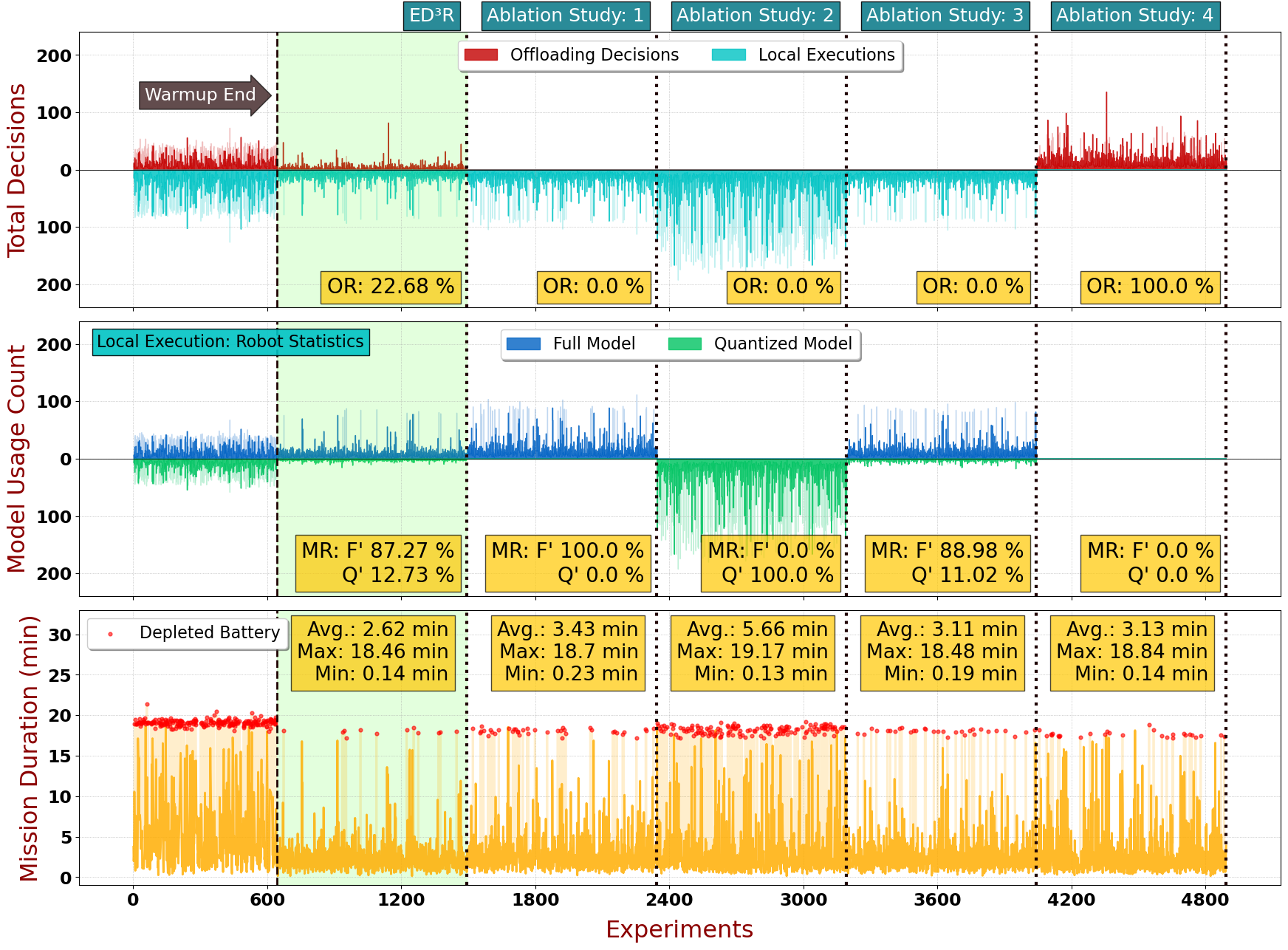}
    \caption{UAV robot's decision-making under 4 ablation studies}
    \label{fig:abl_reward_performance3}
\end{figure}
Figure \ref{fig:abl_reward_performance3} complements the above results as follows:

\noindent\textbf{Figure \ref{fig:abl_reward_performance3} - Remarks:}
\begin{itemize}
    \item The robot's decision-making across ablations is intuitive.
    \item ED$^3$R, once again, achieves on average the lowest mission duration (2.62 minutes), which is especially important in time critical events.
    \item All ablation studies show statistically similar mission duration, except for ablation study 2, which fails to meet performance requirements and results in significantly longer mission duration and as such battery depletions.
\end{itemize}

\textbf{Conclusion 3}: The ablation results show that removing adaptive execution or cooperation leads to inefficient strategies, confirming that hierarchical distributed cooperation is essential.

\textbf{Result 4 - Baseline Study - The superiority of ED$^3$R}: This round of evaluation results highlights the merits of ED$^3$R compared to three baselines: 1) the centralized counterpart of ED$^3$R (ECD$^2$R), 2) an RL approach, namely Independent Proximal Policy Optimization (IPPO), and 3) a distributed rule-based heuristic strategy (RBS). Specifically, IPPO was selected as a multi-agent RL actor-critic that can preserve ED$^3$R's distributed decision-making under the same state and action spaces, mutual reward, mission objectives, and environmental conditions, enabling a direct and fair comparison. The description and configuration of the baselines are given in the Appendix.
Table \ref{tab:baseline_comparisons} shows the consistent superiority of ED$^3$R across both efficiency and mission performance metrics.

\noindent\textbf{Table \ref{tab:baseline_comparisons} - Remarks:}
\begin{itemize}
    \item ED$^3$R achieves the lowest average energy consumption (21.44 kJ), reducing energy by $\sim$16.8 \% compared to ECD$^2$R, $\sim$9.7\% compared to IPPO, and $\sim$38.7\% compared to RBS, while maintaining the highest MSR (97.18\%).
    \item ED$^3$R reduces WER by $\sim$35.6\%, $\sim$31.5\%, and $\sim$62.6\% compared to ECD$^2$R, IPPO and RBS, respectively. This indicates that ED$^3$R enables more effective and reliable exploration with fewer resources.
    \item ECD$^2$R results in higher energy costs due to continuous offloading and centralized processing, while RBS presents a degraded performance due to its heuristic rule-based decision policy.
    \item IPPO is the second-best performing approach with an MSR close to ED$^3$R's, reflecting its ability to effectively adapt to ecosystem's dynamicity through learned neural policies. However, this comes at the cost of higher energy consumption and WER compared to ED$^3$R.
\end{itemize}
  
The performance gap becomes more noticeable under the $85^{th}$ percentile ($P_{85}$), which corresponds to the most demanding missions (top 15\% in terms of total mission duration). ED$^3$R reduces energy consumption by $\sim$30.5\%, $\sim$19.9\% and $\sim$36.4\%, while also lowering WER by $\sim$22.8\%, $\sim$29.6\% and $\sim$15.2\%, compared to ECD$^2$R, IPPO and RBS, respectively. In the case of IPPO, although it achieves the highest MSR, only slightly above ED$^3$R, this comes at a significant cost. ED$^3$R consumes less energy, reduces WER, and completes the mission $\sim$26.4\% faster than IPPO. Most importantly, under such critical events where every minute matters, ED$^3$R manages to reliably detect wildfire events up to $\sim$41\% faster than the baselines.

\textbf{Conclusion 4}: ED$^3$R clearly demonstrates superior performance, showcasing high detection reliability, energy efficiency, robustness, and speed, even under the most demanding scenarios.

\begin{table}[!ht]
\centering
\resizebox{\columnwidth}{!}{%
\begin{tabular}{|
>{\columncolor[HTML]{EFEFEF}}c |
>{\columncolor[HTML]{FFFFFF}}c |
>{\columncolor[HTML]{FFFFFF}}c |
>{\columncolor[HTML]{FFFFFF}}c |
>{\columncolor[HTML]{FFFFFF}}c |
>{\columncolor[HTML]{FFFFFF}}c |
>{\columncolor[HTML]{FFFFFF}}c |}
\hline
\textbf{Strategy} & \cellcolor[HTML]{EFEFEF}\textbf{\begin{tabular}[c]{@{}c@{}}Total \\ Energy\\ (kJ)\end{tabular}} & \cellcolor[HTML]{EFEFEF}\textbf{\begin{tabular}[c]{@{}c@{}}SCS\\ (\%)\end{tabular}} & \cellcolor[HTML]{EFEFEF}\textbf{\begin{tabular}[c]{@{}c@{}}MSR\\ (\%)\end{tabular}} & \cellcolor[HTML]{EFEFEF}\textbf{\begin{tabular}[c]{@{}c@{}}WER \\ (\%)\end{tabular}} & \cellcolor[HTML]{EFEFEF}\textbf{\begin{tabular}[c]{@{}c@{}}LF\\ Accuracy\\ (\%)\end{tabular}} & \cellcolor[HTML]{EFEFEF}\textbf{\begin{tabular}[c]{@{}c@{}}Mission\\ Duration\\ (m)\end{tabular}} \\ \hline
\textbf{ED$^3$R} & \textbf{21.44} & \textbf{90.01} & \textbf{97.18} & \textbf{10.26} & \textbf{74.75} & \textbf{2.61} \\ \hline
\textbf{ECD$^2$R} & 25.76 & 89.54 & 94.59 & 15.94 & 76.72 & 3.31 \\ \hline
\textbf{IPPO} & 23.74 & \multicolumn{1}{l|}{\cellcolor[HTML]{FFFFFF}90.35} & \multicolumn{1}{l|}{\cellcolor[HTML]{FFFFFF}96.73} & \multicolumn{1}{l|}{\cellcolor[HTML]{FFFFFF}14.97} & n/a & 3.03 \\ \hline
\textbf{RBS} & 34.99 & 88.5 & 76.82 & 25.32 & n/a & 4.68 \\ \hline
\cellcolor[HTML]{C0C0C0}\textbf{ED$^3$R ($P_{85}$)} & \textbf{54.34} & \textbf{89.64} & \textbf{76.56} & \textbf{24.51} & \textbf{80.55} & \textbf{6.87} \\ \hline
\cellcolor[HTML]{C0C0C0}\textbf{ECD$^2$R ($P_{85}$)} & 78.18 & 88.53 & 71.88 & 31.74 & \cellcolor[HTML]{FFFFFF}84.18 & 10.33 \\ \hline
\cellcolor[HTML]{C0C0C0}\textbf{IPPO ($P_{85}$)} & 67.86 & 90.12 & 79.52 & 34.8 & n/a & 9.33 \\ \hline
\cellcolor[HTML]{C0C0C0}\textbf{RBS ($P_{85}$)} & 85.41 & 87.73 & 73.44 & 28.92 & \cellcolor[HTML]{FFFFFF}n/a & 11.65 \\ \hline
\end{tabular}%
}
\caption{Comparison of our agentic approach ED$^3$R against three baselines (ECD$^2$R, IPPO \& RBS)}
\label{tab:baseline_comparisons}
\end{table}

\begin{figure}[!ht]
    \centering
    \includegraphics[width=0.95\linewidth]{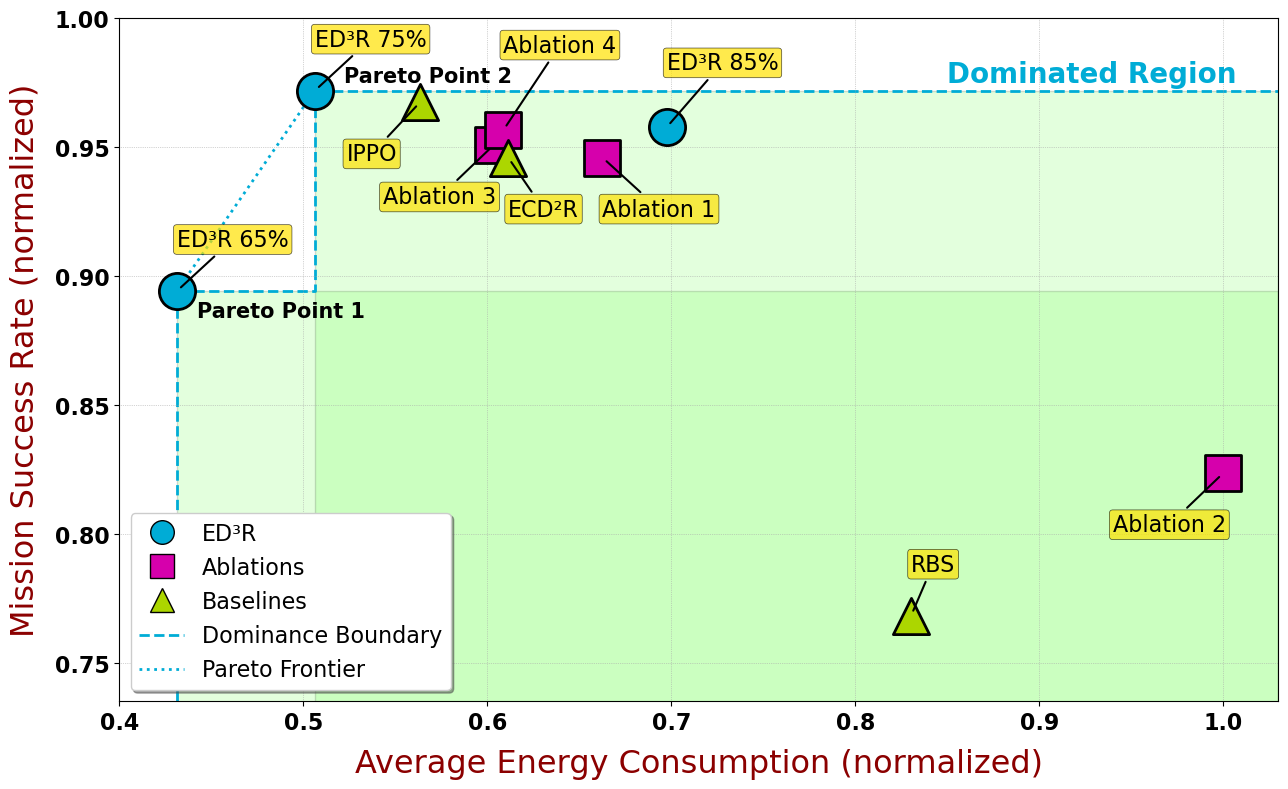}
    \caption{Dominance of ED$^3$R}
    \label{fig:pareto}
\end{figure}

\textbf{Final Result - The dominance of ED$^3$R}: Figure \ref{fig:pareto} captures the essence of this study, focusing on \textit{what matters most}, through a trade-off visualization. The x-axis represents the average normalized energy consumption across missions, while the y-axis captures the mission success rate (MSR). Each point corresponds to a strategy and its trade-off.

We define a strategy as Pareto-optimal if no alternative achieves both lower energy consumption and higher MSR. As such, ED$^3$R dominates as the only strategy that operates along the Pareto frontier. In particular, the 65\% and 75\% configurations define the most effective balance between energy and mission success. Increasing to 85\% yields lower MSR and higher energy, placing it outside the frontier. 

In contrast, all baselines and ablations fall within dominated regions, either consuming more energy for comparable success rates or achieving lower success at similar energy levels, indicating limited trade-off. 

\vspace{-8pt}
\section{Conclusions} \label{conclusions}
This paper introduces ED$^3$R, an energy-aware distributed, hierarchical, and cooperative framework integrating a robot and a remote controller for wildfire detection under \textit{uncertainty}. The controller decides the robot’s motion, while the robot determines \textit{where} and \textit{how} to perform detection. Both agents aim to minimize energy consumption while ensuring detection confidence conformance.

The ED$^3$R is evaluated in 4 rounds. First, 3 different wildfire detection confidence thresholds are examined, showcasing a trade-off between performance requirements and energy efficiency, achieving up to 97.18\% mission success rate. Second, forward-looking accuracy results in more reliable agent decisions and overall mission success. Third, ablation studies show that removing adaptive execution or cooperation leads to less efficient strategies, while ED$^3$R achieves substantial energy savings and higher mission success. Fourth, baseline comparisons demonstrate the superiority of ED$^3$R, especially in demanding scenarios, with an average energy reduction of 21.73\%, up to 41\% faster detections, and up to 20.36\% higher mission success rate.

Finally, a Pareto analysis shows that ED$^3$R is the only strategy operating along the Pareto frontier, providing the most effective balance between energy consumption and mission success. As future work, we will extend our evaluation to real-world experiments and our research toward cooperative robot swarms for monitoring and covering the evolution of hazardous events.

\vspace{-4pt}
\bibliographystyle{IEEEtran}
\bibliography{bibliography/bib}

\begin{thebibliography}{10}
\providecommand{\url}[1]{#1}
\csname url@samestyle\endcsname
\providecommand{\newblock}{\relax}
\providecommand{\bibinfo}[2]{#2}
\providecommand{\BIBentrySTDinterwordspacing}{\spaceskip=0pt\relax}
\providecommand{\BIBentryALTinterwordstretchfactor}{4}
\providecommand{\BIBentryALTinterwordspacing}{\spaceskip=\fontdimen2\font plus
\BIBentryALTinterwordstretchfactor\fontdimen3\font minus \fontdimen4\font\relax}
\providecommand{\BIBforeignlanguage}[2]{{%
\expandafter\ifx\csname l@#1\endcsname\relax
\typeout{** WARNING: IEEEtran.bst: No hyphenation pattern has been}%
\typeout{** loaded for the language `#1'. Using the pattern for}%
\typeout{** the default language instead.}%
\else
\language=\csname l@#1\endcsname
\fi
#2}}
\providecommand{\BIBdecl}{\relax}
\BIBdecl

\bibitem{shukla2025historical}
\BIBentryALTinterwordspacing
G.~Shukla, A.~Kumar, and F.~Parveen, ``A historical perspective and current trends in robotic technology: From mechanization to intelligent automation,'' in \emph{Robot Automation: Principle, Design and Applications}, 1st~ed., R.~Singh, A.~Gehlot, and S.~Maini, Eds.\hskip 1em plus 0.5em minus 0.4em\relax CRC Press, 2025. [Online]. Available: \url{https://www.taylorfrancis.com/chapters/edit/10.1201/9781003560319-1/historical-perspective-current-trends-robotic-technology-geetanjali-shukla-aman-kumar-fraiz-parveen}
\BIBentrySTDinterwordspacing

\bibitem{un2015transforming}
\BIBentryALTinterwordspacing
{United Nations}, ``Transforming our world: The 2030 agenda for sustainable development,'' New York, NY, 2015, united Nations General Assembly Resolution A/RES/70/1. [Online]. Available: \url{https://sdgs.un.org/2030agenda}
\BIBentrySTDinterwordspacing

\bibitem{articleFilho}
W.~Filho, P.~Yang, C.~Li, M.~Cataldi, and I.~Lache, ``Applications of robotics in sustainable development,'' \emph{Sustainable Development}, pp. n/a--n/a, 03 2026.

\bibitem{Ghassemian2026}
M.~Ghassemian, J.~Erfanian, K.~Althoefer, K.~Ahmadi, J.~Eichinger, A.~Meseguer~Valenzuela, P.~Barattini, M.~Bahaei, T.~Booth, A.~Bouttier, C.~Ciochina, M.~Mestoukirdi, P.~Gonçalves, E.~Natalizio, J.~Simonjan, M.~Basaran, F.~Cogen, Y.~Ozsahin, A.~Huete, and T.~Mahmoodi, ``6g architectural foundations and ai-native solutions for future connected robotics.'' 03 2026.

\bibitem{EUSPA2024EMAidUserNeeds}
\BIBentryALTinterwordspacing
``User needs and requirements: Report on emergency management and humanitarian aid,'' European Union Agency for the Space Programme (EUSPA), Tech. Rep., 2024. [Online]. Available: \url{www.euspa.europa.eu/publications-multimedia/publications/user-needs-and-requirements}
\BIBentrySTDinterwordspacing

\bibitem{one6G2023}
\BIBentryALTinterwordspacing
``6g and robotics: A methodology to identify potential service requirements for 6g-empowered robotic use cases,'' one6G Association, Tech. Rep., November 2023. [Online]. Available: \url{https://one6g.org/one6g-publishes-white-paper-on-6g-service-needs-in-the-domain-of-robotics/}
\BIBentrySTDinterwordspacing

\bibitem{6G-IA_2024_EuropeanVision}
\BIBentryALTinterwordspacing
{6G-IA Vision Working Group}, ``European vision for the 6g network ecosystem,'' 6G-IA, Tech. Rep. Version 2.0, 2024. [Online]. Available: \url{https://6g-ia.eu/wp-content/uploads/2024/11/european-vision-for-the-6g-network-ecosystem.pdf}
\BIBentrySTDinterwordspacing

\bibitem{10142027Romero}
A.~Romero, C.~Delgado, L.~Zanzi, X.~Li, and X.~Costa-Pérez, ``Oros: Online operation and orchestration of collaborative robots using 5g,'' \emph{IEEE Transactions on Network and Service Management}, vol.~20, no.~4, pp. 4216--4230, 2023.

\bibitem{9762674Yin}
R.~Yin, Y.~Shen, H.~Zhu, X.~Chen, and C.~Wu, ``Time-critical tasks implementation in mec based multi-robot cooperation systems,'' \emph{China Communications}, vol.~19, no.~4, pp. 199--215, 2022.

\bibitem{drones8100564}
\BIBentryALTinterwordspacing
Y.~Lu, C.~Xu, and Y.~Wang, ``Joint computation offloading and trajectory optimization for edge computing uav: A knn-ddpg algorithm,'' \emph{Drones}, vol.~8, no.~10, 2024. [Online]. Available: \url{https://www.mdpi.com/2504-446X/8/10/564}
\BIBentrySTDinterwordspacing

\bibitem{Djenadi02012022}
\BIBentryALTinterwordspacing
A.~Djenadi and B.~Mendil, ``Energy-aware task allocation strategy for multi robot system,'' \emph{International Journal of Modelling and Simulation}, vol.~42, no.~1, pp. 153--167, 2022. [Online]. Available: \url{https://doi.org/10.1080/02286203.2020.1855405}
\BIBentrySTDinterwordspacing

\bibitem{XUE2026109171}
\BIBentryALTinterwordspacing
Y.~Xue, C.~K. Tan, and W.~P. Wong, ``Energy-aware multi-robot exploration and coverage in fragmented unknown environments using collaborative reinforcement learning,'' \emph{Results in Engineering}, vol.~29, p. 109171, 2026. [Online]. Available: \url{https://www.sciencedirect.com/science/article/pii/S2590123026002148}
\BIBentrySTDinterwordspacing

\bibitem{Han8040138}
\BIBentryALTinterwordspacing
D.~Han, H.~Jiang, L.~Wang, X.~Zhu, Y.~Chen, and Q.~Yu, ``Collaborative task allocation and optimization solution for unmanned aerial vehicles in search and rescue,'' \emph{Drones}, vol.~8, no.~4, 2024. [Online]. Available: \url{https://www.mdpi.com/2504-446X/8/4/138}
\BIBentrySTDinterwordspacing

\bibitem{Farsath10580372}
K.~Rashida~Farsath, K.~Jitha, V.~Mohammed~Marwan, A.~Muhammed Ali~Jouhar, K.~Muhammed~Farseen, and K.~Musrifa, ``Ai-enhanced unmanned aerial vehicles for search and rescue operations,'' in \emph{2024 5th International Conference on Innovative Trends in Information Technology (ICITIIT)}, 2024, pp. 1--10.

\bibitem{Horyna101007}
\BIBentryALTinterwordspacing
J.~Horyna, T.~Baca, V.~Walter, D.~Albani, D.~Hert, E.~Ferrante, and M.~Saska, ``Decentralized swarms of unmanned aerial vehicles for search and rescue operations without explicit communication,'' \emph{Auton. Robots}, vol.~47, no.~1, p. 77–93, Oct. 2022. [Online]. Available: \url{https://doi.org/10.1007/s10514-022-10066-5}
\BIBentrySTDinterwordspacing

\bibitem{XING2022102972}
\BIBentryALTinterwordspacing
L.~Xing, X.~Fan, Y.~Dong, Z.~Xiong, L.~Xing, Y.~Yang, H.~Bai, and C.~Zhou, ``Multi-uav cooperative system for search and rescue based on yolov5,'' \emph{International Journal of Disaster Risk Reduction}, vol.~76, p. 102972, 2022. [Online]. Available: \url{https://www.sciencedirect.com/science/article/pii/S2212420922001911}
\BIBentrySTDinterwordspacing

\bibitem{Urbaniak2024}
D.~Urbaniak, S.~Damsgaard, W.~Zhang, J.~Rosell, R.~Suarez, and M.~Suppa, ``Distributed control for collaborative robotic systems using 5g edge computing,'' \emph{IEEE Access}, vol.~PP, pp. 1--1, 01 2024.

\bibitem{9676458Zhou}
Y.~Zhou, J.~Xiao, Y.~Zhou, and G.~Loianno, ``Multi-robot collaborative perception with graph neural networks,'' \emph{IEEE Robotics and Automation Letters}, vol.~7, no.~2, pp. 2289--2296, 2022.

\bibitem{damigos2024}
\BIBentryALTinterwordspacing
G.~Damigos, A.~Saradagi, S.~Sandberg, and G.~Nikolakopoulos, ``Environmental awareness dynamic 5g qos for retaining real time constraints in robotic applications,'' 2024. [Online]. Available: \url{https://arxiv.org/abs/2402.06470}
\BIBentrySTDinterwordspacing

\bibitem{10002890Cui}
Q.~Cui, X.~Zhao, W.~Ni, Z.~Hu, X.~Tao, and P.~Zhang, ``Multi-agent deep reinforcement learning-based interdependent computing for mobile edge computing-assisted robot teams,'' \emph{IEEE Transactions on Vehicular Technology}, vol.~72, no.~5, pp. 6599--6610, 2023.

\bibitem{8331947Pham}
H.~X. Pham, H.~M. La, D.~Feil-Seifer, and M.~C. Deans, ``A distributed control framework of multiple unmanned aerial vehicles for dynamic wildfire tracking,'' \emph{IEEE Transactions on Systems, Man, and Cybernetics: Systems}, vol.~50, no.~4, pp. 1537--1548, 2020.

\bibitem{9504947Shrestha}
K.~Shrestha, R.~Dubey, A.~Singandhupe, S.~Louis, and H.~La, ``Multi objective uav network deployment for dynamic fire coverage,'' in \emph{2021 IEEE Congress on Evolutionary Computation (CEC)}, 2021, pp. 1280--1287.

\bibitem{Patrinopoulou2024}
N.~Patrinopoulou, I.~Daramouskas, D.~Meimetis, V.~Lappas, and V.~Kostopoulos, ``A distributed framework for persistent wildfire monitoring with fixed wing uavs,'' \emph{Drones and Autonomous Vehicles}, vol.~1, no.~3, p. 10009, 2024.

\bibitem{Julian2019}
\BIBentryALTinterwordspacing
K.~D. Julian and M.~J. Kochenderfer, ``Distributed wildfire surveillance with autonomous aircraft using deep reinforcement learning,'' 2018. [Online]. Available: \url{https://arxiv.org/abs/1810.04244}
\BIBentrySTDinterwordspacing

\bibitem{Seraj2022}
\BIBentryALTinterwordspacing
E.~Seraj, A.~Silva, and M.~Gombolay, ``Multi-uav planning for cooperative wildfire coverage and tracking with quality-of-service guarantees,'' \emph{Autonomous Agents and Multi-Agent Systems}, vol.~36, no.~2, p.~39, 2022. [Online]. Available: \url{https://doi.org/10.1007/s10458-022-09566-6}
\BIBentrySTDinterwordspacing

\bibitem{11315192Akpomedaye}
B.~Akpomedaye, A.~Shalan, N.~A. Walee, and M.~Rahman, ``Energy-efficient uav surveillance for early wildfire detection using ai-driven image analysis,'' in \emph{2025 IEEE/ACS 22nd International Conference on Computer Systems and Applications (AICCSA)}, 2025, pp. 1--7.

\bibitem{10206033Suo}
J.~Suo, X.~Zhang, W.~Shi, and W.~Zhou, ``E3-uav: An edge-based energy-efficient object detection system for unmanned aerial vehicles,'' \emph{IEEE Internet of Things Journal}, vol.~11, no.~3, pp. 4398--4413, 2024.

\bibitem{khoshdel2024pyrotrack}
\BIBentryALTinterwordspacing
S.~Khoshdel, Q.~Luo, and F.~Afghah, ``Pyrotrack: Belief-based deep reinforcement learning path planning for aerial wildfire monitoring in partially observable environments,'' 2024. [Online]. Available: \url{https://arxiv.org/abs/2403.11095}
\BIBentrySTDinterwordspacing

\bibitem{drones9010017}
\BIBentryALTinterwordspacing
P.~Zhu, R.~Song, J.~Zhang, Z.~Xu, Y.~Gou, Z.~Sun, and Q.~Shao, ``Multiple uav swarms collaborative firefighting strategy considering forest fire spread and resource constraints,'' \emph{Drones}, vol.~9, no.~1, 2025. [Online]. Available: \url{https://www.mdpi.com/2504-446X/9/1/17}
\BIBentrySTDinterwordspacing

\bibitem{11005691Diaz}
C.~Diaz-Vilor, M.~Barzegaran, and H.~Jafarkhani, ``Multi-uav energy-efficient wildfire coverage optimization,'' \emph{IEEE Transactions on Wireless Communications}, vol.~24, no.~10, pp. 8633--8648, 2025.

\bibitem{10834526Diaz}
C.~Diaz-Vilor, A.~Lozano, and H.~Jafarkhani, ``A reinforcement learning approach for wildfire tracking with uav swarms,'' \emph{IEEE Transactions on Wireless Communications}, vol.~24, no.~4, pp. 2766--2782, 2025.

\bibitem{10838709Mu}
L.~Mu, Y.~Yang, B.~Wang, Y.~Zhang, N.~Feng, and X.~Xie, ``Edge computing-based real-time forest fire detection using uav thermal and color images,'' \emph{IEEE Journal of Selected Topics in Applied Earth Observations and Remote Sensing}, vol.~18, pp. 6760--6771, 2025.

\bibitem{Thibbotuwawa101007}
A.~Thibbotuwawa, P.~Nielsen, B.~Zbigniew, and G.~Bocewicz, ``Energy consumption in unmanned aerial vehicles: A review of energy consumption models and their relation to the uav routing,'' in \emph{Information Systems Architecture and Technology: Proceedings of 39th International Conference on Information Systems Architecture and Technology -- ISAT 2018}, J.~Swiatek, L.~Borzemski, and Z.~Wilimowska, Eds.\hskip 1em plus 0.5em minus 0.4em\relax Cham: Springer International Publishing, 2019, pp. 173--184.

\bibitem{Pollet03832135}
\BIBentryALTinterwordspacing
F.~Pollet, S.~Delbecq, M.~Budinger, and J.-M. Moschetta, ``{Design optimization of multirotor drones in forward flight},'' in \emph{{32nd Congress of the International Council of the Aeronautical Sciences}}, Shanghai, China, Sep. 2021. [Online]. Available: \url{https://hal.science/hal-03832135}
\BIBentrySTDinterwordspacing

\bibitem{Dorling7513397}
K.~Dorling, J.~Heinrichs, G.~G. Messier, and S.~Magierowski, ``Vehicle routing problems for drone delivery,'' \emph{IEEE Transactions on Systems, Man, and Cybernetics: Systems}, vol.~47, no.~1, pp. 70--85, 2017.

\bibitem{Marins8691921}
J.~L. Marins, T.~M. Cabreira, K.~S. Kappel, and P.~R. Ferreira, ``A closed-form energy model for multi-rotors based on the dynamic of the movement,'' in \emph{2018 VIII Brazilian Symposium on Computing Systems Engineering (SBESC)}, 2018, pp. 256--261.

\bibitem{Muli101007}
C.~Muli, S.~Park, and M.~Liu, ``A comparative study on energy consumption models for drones,'' in \emph{Internet of Things}, A.~Gonz{\'a}lez-Vidal, A.~Mohamed~Abdelgawad, E.~Sabir, S.~Ziegler, and L.~Ladid, Eds.\hskip 1em plus 0.5em minus 0.4em\relax Cham: Springer International Publishing, 2022, pp. 199--210.

\bibitem{Luo2024}
\BIBentryALTinterwordspacing
X.~Luo, A.~Rechardt, G.~Sun, K.~K. Nejad, F.~Yáñez, B.~Yilmaz, K.~Lee, A.~O. Cohen, V.~Borghesani, A.~Pashkov, D.~Marinazzo, J.~Nicholas, A.~Salatiello, I.~Sucholutsky, P.~Minervini, S.~Razavi, R.~Rocca, E.~Yusifov, T.~Okalova, N.~Gu, M.~Ferianc, M.~Khona, K.~R. Patil, P.-S. Lee, R.~Mata, N.~E. Myers, J.~K. Bizley, S.~Musslick, I.~P. Bilgin, G.~Niso, J.~M. Ales, M.~Gaebler, N.~A.~R. Murty, L.~Loued-Khenissi, A.~Behler, C.~M. Hall, J.~Dafflon, S.~D. Bao, and B.~C. Love, ``Large language models surpass human experts in predicting neuroscience results,'' \emph{Nature Human Behaviour}, 2024. [Online]. Available: \url{https://doi.org/10.1038/s41562-024-02046-9}
\BIBentrySTDinterwordspacing

\bibitem{gazebo}
\BIBentryALTinterwordspacing
Official gazebo website. [Online]. Available: \url{https://gazebosim.org}
\BIBentrySTDinterwordspacing

\bibitem{ros2}
\BIBentryALTinterwordspacing
Official ros website. [Online]. Available: \url{https://www.ros.org/}
\BIBentrySTDinterwordspacing

\bibitem{ode}
\BIBentryALTinterwordspacing
Open dynamics engine. [Online]. Available: \url{https://www.ode.org/}
\BIBentrySTDinterwordspacing

\bibitem{blender}
\BIBentryALTinterwordspacing
Blender free \& open source 3d software. [Online]. Available: \url{https://www.blender.org/download/}
\BIBentrySTDinterwordspacing

\bibitem{iris_drone}
\BIBentryALTinterwordspacing
3dr iris quadcopter specifications. [Online]. Available: \url{https://www.arducopter.co.uk/iris-quadcopter-uav.html}
\BIBentrySTDinterwordspacing

\bibitem{dfire_git}
\BIBentryALTinterwordspacing
Dfire dataset official github. [Online]. Available: \url{https://github.com/gaiasd/DFireDataset}
\BIBentrySTDinterwordspacing

\bibitem{dfire}
\BIBentryALTinterwordspacing
P.~V. A.~B. de~Venâncio, R.~J. Campos, T.~M. Rezende, A.~C. Lisboa, and A.~V. Barbosa, ``A hybrid method for fire detection based on spatial and temporal patterns,'' \emph{Neural Computing and Applications}, vol.~35, no.~13, pp. 9349--9361, 2023. [Online]. Available: \url{https://doi.org/10.1007/s00521-023-08260-2}
\BIBentrySTDinterwordspacing

\bibitem{m4sfwd_git}
\BIBentryALTinterwordspacing
M4sfwd dataset official github. [Online]. Available: \url{https://github.com/Philharmy-Wang/M4SFWD}
\BIBentrySTDinterwordspacing

\bibitem{m4sfwd}
\BIBentryALTinterwordspacing
G.~Wang, ``Multiple scenarios, multiple weather conditions, multiple lighting conditions and multiple wildfire objects synthetic forest wildfire dataset (m4sfwd),'' 2024. [Online]. Available: \url{https://dx.doi.org/10.21227/m9kz-bw61}
\BIBentrySTDinterwordspacing

\end{thebibliography}

\appendix \label{appendix}

\subsection{Power for Movement Energy Estimation}

\textbf{Hovering Power}: The power required by the robot to counteract gravitational forces and maintain altitude using its propellers, denoted by $P_{\mathrm{hov}}$, is:
    \begin{equation}
        P_{\mathrm{hov}} = m^\frac{3}{2} \cdot \sqrt{\frac{g^3}{2 \cdot n^{\mathrm{prop}}} \cdot \rho \cdot A^{\mathrm{prop}}} \quad (Watts),
    \end{equation}
where $m$, $n^{\mathrm{prop}}$, $A^{\mathrm{prop}}$ are the total mass ($kg$), number of propellers and the area of each propeller ($m^2$) of the robot, respectively. $g$ denotes the gravitational acceleration ($m/s^2$) and $\rho$ is the air density ($kg/m^3$).
     
\textbf{Horizontal Power}: The power required by the robot to overcome gravitational and aerodynamic drag forces in the horizontal direction to move forward, backward, left, or right, denoted by $P_{\mathrm{hor}}$, is:
\begin{equation}
    P_{\mathrm{hor}} = P_{\mathrm{hov}} + \frac{1}{2} \cdot \rho \cdot A \cdot (v^\mathtt{h})^3 \cdot C^{\mathtt{dh}} \; (Watts),
\end{equation}
where $A$ is the cross-sectional area of the robot facing the direction of motion ($m^2$), $v^\mathtt{h}$ is the horizontal speed ($m/s$), and $C^{\mathtt{dh}}$ is the horizontal drag coefficient.
    
\textbf{Vertical Power}: The power required by the robot to overcome gravitational and aerodynamic drag forces in the vertical direction to move upward or downward, denoted by $P_{\mathrm{ver}}$, is:
    \begin{equation}
        P_{\mathrm{ver}} = m \cdot g  + \frac{1}{2} \cdot \rho \cdot A \cdot (v^\mathtt{v})^3 \cdot C^{\mathtt{dv}} \cdot \mathbb{I}_{v^\mathtt{v}}\quad (Watts),
    \end{equation}
    where:
    \vspace{-5pt}
    \begin{equation}
        \mathbb{I}_{v^\mathtt{v}} = 
        \begin{cases}
          1 & \text{if $v^\mathtt{v} \geq 0$ \quad(upwards)} \\
          -1 & \text{if $v^\mathtt{v} < 0$ \quad(downwards)}
        \end{cases},
    \end{equation}
$v^\mathtt{v}$, and $C^{\mathtt{dv}}$ are the vertical speed ($m/s$) and vertical drag coefficient of the robot, respectively. $\mathbb{I}_{v^\mathtt{v}}$ is an indicator function that adjusts the sign of the drag term based on the direction of vertical motion.

\subsection{Achievable Data Rate Modeling}
The achievable data rate $\mathtt{d}_{k}$ is given by:
\vspace{-4pt}
\begin{equation}
  \mathtt{d}_{k} = {b_{k} \cdot log_2\left(1 + \frac{g_{k} \cdot p_{k}}{b_{k} \cdot N_0}\right)},
\end{equation}
where: \vspace{-4pt}
\begin{equation}
  g_{k} = 10 ^{-\frac{\mathtt{L}_0 + 10 \cdot \mathtt{e} \cdot \log_{10}{(\frac{\delta_{k,\text{RC}}}{\mathtt{\delta_0}}})}{10}},
\end{equation}
\vspace{-8pt}
\begin{equation}
    \mathtt{L}_0 = 20 \cdot \log_{10}(\mathtt{\delta}_0) + 20 \cdot \log_{10}(\mathtt{f}_{k})  +20 \cdot \log_{10}(\frac{4 \cdot \pi}{\mathtt{c}}),
\end{equation}
\vspace{-8pt}
\begin{equation}
  p_{k} = tp_{k} + \mathtt{L_0}.
\end{equation}
$tp_{k}$ is the target signal strength ($dBm$) $\mathtt{L}_0$ is the path loss ($dB$) at the reference distance $\mathtt{\delta}_0$ ($meters$), $\delta_{k,\text{RC}}$ is the distance between the robot and the $\text{RC}$ at $k$ ($meters$), $\mathtt{f}_{k}$ denotes the transmission carrier frequency ($Hz$), $\mathtt{e}$ is the path loss exponent and $\mathtt{c}$ the speed of light ($m/s$).
\subsection{UAV Robot Specifications}
Table \ref{tab:robot_setup} summarizes the UAV robot's configuration. Several abbreviations are used in the table: Frames Per Second (FPS), Width ($\times$) Height (WxH), Height/Width/Depth (H/W/D), and Horizontal/Vertical (H/V).

\begin{table}[h]
\centering
\begin{subtable}[t]{0.32\textwidth}
\centering
\resizebox{\textwidth}{!}{%
\begin{tabular}{|>{\columncolor[HTML]{EFEFEF}}l |l|}
\hline
\multicolumn{1}{|c|}{\cellcolor[HTML]{EFEFEF}\textbf{Parameter}} &
\multicolumn{1}{c|}{\cellcolor[HTML]{EFEFEF}\textbf{Value}} \\ \hline
\textbf{\# of LiDARs} & 6 \\ \hline
\textbf{LiDAR Safety Distance} ($\mathtt{sd}_{\mathrm{min}}$) & 2 m \\ \hline
\textbf{LiDAR Sensor Power} ($P_{\mathrm{LiDAR}}$) & 0.9 W \\ \hline
\textbf{Camera Sensor Power} ($P_{\mathrm{camera}}$) & 2 W \\ \hline
\textbf{Camera FPS} ($I_{\mathrm{camera}}$) & 11 \\ \hline
\textbf{Camera Resolution (WxH)} & 1920 x 1080 \\ \hline
\textbf{Camera Frame Size} ($w_{\mathrm{camera}}$) & 519 KByte \\ \hline
\textbf{Sensing Period ($\mathtt{s}$)} & 3 s \\ \hline
\textbf{Gravitational Acceleration ($g$)} & 9.8 m/s$^2$ \\ \hline
\end{tabular}%
}
\caption{Sensing \& Perception}
\label{tab:robot_sensing}
\end{subtable}
\hfill
\begin{subtable}[t]{0.32\textwidth}
\centering
\resizebox{\textwidth}{!}{%
\begin{tabular}{|>{\columncolor[HTML]{EFEFEF}}l |l|}
\hline
\multicolumn{1}{|c|}{\cellcolor[HTML]{EFEFEF}\textbf{Parameter}} &
\multicolumn{1}{c|}{\cellcolor[HTML]{EFEFEF}\textbf{Value}} \\ \hline
\textbf{Battery Capacity} ($\beta^{\mathrm{max}}$) & 3500 mAh \\ \hline
\textbf{CPU Speed} ($f^{\mathrm{min}}/f^{\mathrm{max}}$) & 1.5/3.2 GHz \\ \hline
\textbf{CPU Cores} ($n^{\mathrm{cpu}}$) & 6 \\ \hline
\textbf{CPU ESC} ($\varsigma$) & 1e-28 \\ \hline
\textbf{CPU FLOPs/Cycle} ($c$) & 8 \\ \hline
\textbf{Transmission Power} ($p^{\mathrm{min}}/p^{\mathrm{max}}$) & 14/33 dBm \\ \hline
\end{tabular}%
}
\caption{Computation \& Communication}
\label{tab:robot_compute}
\end{subtable}
\hfill
\begin{subtable}[t]{0.32\textwidth}
\centering
\resizebox{\textwidth}{!}{%
\begin{tabular}{|>{\columncolor[HTML]{EFEFEF}}l |l|}
\hline
\multicolumn{1}{|c|}{\cellcolor[HTML]{EFEFEF}\textbf{Parameter}} &
\multicolumn{1}{c|}{\cellcolor[HTML]{EFEFEF}\textbf{Value}} \\ \hline
\textbf{Mass ($m$)} & 1.816 kg \\ \hline
\textbf{\# of Propellers ($n^{\mathrm{prop}}$)} & 4 \\ \hline
\textbf{Propeller Radius ($A^{\mathrm{prop}}$)} & 100 mm \\ \hline
\textbf{UAV Size (H/W/D)} & 230/470/470 mm \\ \hline
\textbf{H/V Speed ($v^\mathtt{v}/v^\mathtt{h}$)} & 5 m/s \\ \hline
\textbf{H/V Drag Coeff. ($C^{\mathtt{dv}}/C^{\mathtt{dh}}$)} & 0.35/0.55 \\ \hline
\end{tabular}%
}
\caption{Physical Specifications}
\label{tab:robot_physical}
\end{subtable}

\caption{UAV Robot Configuration Setup}
\label{tab:robot_setup}
\end{table}

\subsection{Computer Vision Detection Model Configuration}

The Full and Quantized CV models were trained offline on publicly available datasets, namely DFire \cite{dfire_git,dfire} and M$^4$SFWD \cite{m4sfwd_git, m4sfwd}, which together contain approximately 25K diverse images. Their configurations are summarized in Table \ref{tab:cv_models}.

\begin{table}[!ht]
\centering
\resizebox{200pt}{!}{%
\begin{tabular}{|p{0.34\linewidth}|p{0.24\linewidth}|p{0.24\linewidth}|}
\hline
\cellcolor[HTML]{EFEFEF}\textbf{Configuration} & \cellcolor[HTML]{EFEFEF}\textbf{Full CNN} & \cellcolor[HTML]{EFEFEF}\textbf{Quantized CNN} \\ \hline
\cellcolor[HTML]{EFEFEF}\textbf{\# of Layers} & \multicolumn{2}{c|}{11 convolutional + 2 fully connected} \\ \hline
\cellcolor[HTML]{EFEFEF}\textbf{Activation Function} & \multicolumn{2}{c|}{LeakyReLU} \\ \hline
\cellcolor[HTML]{EFEFEF}\textbf{Learning Rate} & \multicolumn{2}{c|}{\(2 \times 10^{-4}\)} \\ \hline
\cellcolor[HTML]{EFEFEF}\textbf{Precision} & \texttt{float32} & \texttt{qint8} \\ \hline
\cellcolor[HTML]{EFEFEF}\textbf{Trainable Parameters} & \multicolumn{2}{c|}{$36.5 \text{M}$ }\\ \hline
\cellcolor[HTML]{EFEFEF}\textbf{Complexity (GFLOPs)} & 20.23 & 5.05 \\ \hline
\end{tabular}
}
\caption{CV models used for wildfire event detection}
\label{tab:cv_models}
\end{table}

\subsection{Wireless Communication Network Specifications}
Table \ref{tab:network_setup} provides the configuration of the network model.

\begin{table}[!ht]
\centering
\resizebox{200pt}{!}{%
\begin{tabular}{|
>{\columncolor[HTML]{EFEFEF}}l |l|}
\hline
\multicolumn{1}{|c|}{\cellcolor[HTML]{EFEFEF}\textbf{Parameter}} & \multicolumn{1}{c|}{\cellcolor[HTML]{EFEFEF}\textbf{Value}} \\ \hline
\textbf{Target Signal Strength} ($tp_{k}$) & -80 dBm \\ \hline
\textbf{$\text{RC}$ Antenna Frequency} ($\mathtt{f}_{k}$) & 2.4 GHz \\ \hline
\textbf{Reference Distance} ($\mathtt{\delta_0}$) & 1 m \\ \hline
\textbf{Bandwidth min/max} ($b_{\mathrm{min}}/b_{\mathrm{max}}$) & 15/30 MHz \\ \hline
\textbf{Spectral Efficiency min/max }($se_{\mathrm{min}}/se_{\mathrm{max}}$) & 0.12/4 \\ \hline
\textbf{Noise Spectral Density} ($N_0$) & -158 dBm/Hz \\ \hline
\textbf{Path Loss Exponent} ($\mathtt{e}$) & 3.8 \\ \hline
\textbf{Shadow Fading} & 8 dB \\ \hline
\end{tabular}%
}
\caption{Network Configuration Setup}
\label{tab:network_setup}
\vspace{-8pt}
\end{table}

\subsection{Baselines}
\textbf{Centralized Counterpart of ED$^3$R (ECD$^2$R)}:
In ECD$^2$R the $\text{RC}$ is responsible for both the motion selection and CV detection execution, while the UAV robot acts only as a sensing device. The $\text{RC}$ RFNN (Table \ref{tab:DSAR-NN-Configuration}) is retrained under this configuration. 

\textbf{Distributed Rule-based Heuristic Strategy (RBS)}: 
RBS performs a random-walk inspection at each timestep and determines whether to execute the task locally or offload it by comparing the transmission ratio ${\mathrm{ratio}}^{(\mathrm{transmission})}_{k}=\frac{\mathtt{d}_{k}}{d_{k}}$ and the processing ratio ${\mathrm{ratio}}^{(\mathrm{processing})}_{k}=\frac{f_{k}}{f^{\max}}$ against predefined thresholds $\mathrm{thr}_1=0.85$ and $\mathrm{thr}_2=0.65$. RBS selects the strategy whose corresponding ratio exceeds its threshold. If both exceed, the choice is random, while if neither does, the strategy with the minimum distance from its respective threshold is selected.
RBS integrates the early mission completion mechanism defined in ED$^3$R to adapt exploration based on detection performance.\\
\textbf{Independent Proximal Policy Optimization (IPPO)}: Two independent PPO RL agents are considered, one for the RC and one for the robot. Both the RC and robot agents employ separate actor and critic MultiLayer Perceptrons (MLPs) with two fully connected hidden layers of 256 neurons and $\tanh$ activations. A batch size of 256, the Adam optimizer with learning rate 3e-04, a discount factor of 0.99, minibatches of 64, and an entropy coefficient of 0.01 were used. IPPO was trained using the same number of missions as ED$^3$R and the selected hyperparameters remained fixed throughout the experiments. 

\end{document}